\documentclass[11pt]{article}

\usepackage[final]{acl}

\usepackage{times}
\usepackage{latexsym}
\usepackage{graphicx}
\usepackage{subcaption}
\usepackage{booktabs}
\usepackage{amsmath}
\usepackage{amssymb}
\usepackage{caption}
\usepackage{array}
\usepackage{multirow}
\usepackage{verbatim}
\usepackage{microtype}
\usepackage{float}
\usepackage{tikz}

\usetikzlibrary{arrows.meta, positioning, fit, backgrounds, calc, shapes.geometric}
\usepackage{pgfplots}
\pgfplotsset{compat=1.18}

\definecolor{barblue}{RGB}{95,161,195}
\definecolor{barred}{RGB}{218,98,71}
\usepackage[T1]{fontenc}

\usepackage[utf8]{inputenc}

\usepackage{microtype}

\usepackage{inconsolata}

\usepackage{graphicx}

\title{An Interpretable Linguistically-Grounded Analysis of Scene Saliency in Movie Screenplays}

\author{
{\normalfont \textbf{Uday Bindal}}$^{*}$ \hspace{1.2em} {\normalfont \textbf{Ishaan Karan}}$^{*}$ \hspace{1.2em} {\normalfont \textbf{Manan Garg}}$^{*}$ \\[4pt]
\textbf{Ekansh Goyal} \hspace{1.2em} \textbf{Rahul Mishra} \hspace{1.2em} \textbf{Rajakrishnan Rajkumar} \\[6pt]
IIIT Hyderabad \\[4pt]
\texttt{\{uday.bindal,ishaan.karan,ekansh.goyal\}@research.iiit.ac.in} \\
\texttt{manan.garg@students.iiit.ac.in} \\
\texttt{\{rahul.mishra,raja\}@iiit.ac.in}
}

\begin{document}
\maketitle


\renewcommand{\thefootnote}{}
\footnotetext[1]{\textsuperscript{*}Equal contribution. \quad \textsuperscript{1}Code: \url{https://github.com/Ishaank1805/lgsm-screenplay-salience} \quad Features: \url{https://huggingface.co/datasets/Ishaank18/screenplay-features-linguistic}}
\renewcommand{\thefootnote}{\arabic{footnote}}

\begin{abstract}

We investigate the factors affecting salience in movie screenplay by posing two research questions: 1. {\em Does Uniform Information Density (UID) predict salient scenes?} 2. {\em What other linguistic properties distinguish salient scenes from non-salient background?} We answer both questions through a controlled corpus study on the MENSA corpus, paired with a Linguistically-Grounded Salience Model (LGSM), a dual-stream classifier that fuses a frozen BERT encoder with explicit linguistic features through adaptive gating. The UID picture is layered in long-form discourse. Per-scene UID summaries carry no independent signal once length is controlled, but at the token level salient scenes match non-salient material in the bulk while specifically suppressing the rarest, hardest-to-predict tokens in the upper tail. UID therefore lives at the within-scene token level rather than as a per-scene aggregate. Salient scenes also show a moderate and sustained cast of central characters (following an inverted-U pattern), strong third-person usage, syntactically simpler language, and more abstract vocabulary about motivation and conflict. LGSM exploits this signature and improves recall on salient scenes over a strong semantic baseline while yielding better downstream summaries. 

\end{abstract}

\vspace{-10pt}
\section{Introduction}\label{sect:intro}
The concept of salience has been explored extensively across psycholinguistics, visual cognition
and Natural Language Processing (NLP). Salience is defined as the degree to which a stimulus at-
tracts attention within a specific setting~\citep{awh2012}. The relationship between salience and predictability is not straightforward: both highly predictable and highly surprising elements can be salient~\citep{SchmidGunther2016UnifiedFramework,salience-zarcone2016}. The Uniform Information Density (UID) hypothesis states that speakers spread information evenly through their utterances, avoiding sharp spikes and troughs~\citep{fenk1980,levy2007speakers,jaeger2010redundancy}. UID has been supported at the {\bf sentence level} for lexical choices~\citep{levy2007speakers,jaeger2010redundancy} and word order~\citep{uid-word-order-2023}, but its role in long-form narrative is largely untested~\citep{tsipidi2024surprise}. In this paper, extending prior works at the sentence level cited above, we pursue two questions pertaining to salience in longform discourse ({\em viz.}, movie screenplays):

\hspace{-4mm}\textbf{(1)} {\em Does UID predict salient scenes?} \\
\textbf{(2)} {\em What other linguistic properties distinguish salient scenes from non-salient ones?} 

Salience detection in longform fiction is harder than in news. News articles follow an inverted-pyramid convention with key content at the start, exploited by lead-bias~\citep{See2017} and centrality~\citep{Mihalcea2004} priors. Long-form fiction has no such prior: salient information is spread across thousands of tokens, with a climactic concentration rather than an early one~\citep{papalampidi2019movie}. News-style heuristics therefore fail on screenplays, and explicit structural cues alongside neural representations appear to help~\citep{saxena2024select}.

We use two complementary methods to investigate our research questions: 1. A controlled corpus study (\S\ref{sec:corpus_study}) on the MENSA dataset of movie screenplays~\citep{saxena2024select} uses post-LASSO logistic regression~\citep{tibshirani1996regression,belloni2013least} to identify significant linguistic predictors of salience. 2. We implement the Linguistically-Grounded Salience Model (LGSM, \S\ref{sect:methods}), which combines a frozen BERT encoder~\citep{devlin-etal-2019-bert} with an explicit linguistic-feature stream through Adaptive Gated Fusion. 
\footnotemark[1] We then use SHAP attribution~\citep{Lundberg2017} on the trained classifier to identify which feature families it relies on.

LGSM lifts recall on the salient class from 51.56\% to 68.34\% over a semantic baseline and improves downstream summarisation by +3.67 ROUGE-1 with a stable BERTScore (83.17 vs.\ 83.29). Five findings emerge: three positive (identifying signals that carry independent salience information), and two negative (constraining how the problem should be modelled). \textbf{(P1)} A token-level upper-tail surprisal asymmetry: salient and non-salient scenes share nearly identical word-level surprisal distributions in the bulk, but salient scenes suppress the rarest, hardest-to-predict tokens in the upper tail. \textbf{(P2)} A recurring linguistic signature: an intermediate number of central characters (inverted-U), sustained third-person pronominal reference, syntactic simplicity, and elevated abstract language. \textbf{(P3)} The LGSM architecture itself, whose recall and summarisation gains are reported above. \textbf{(N1)} Scene-aggregate UID summaries carry no independent signal once length and positional covariates are controlled. \textbf{(N2)} RST and semantic-grounding features act as precision filters, not recall boosters. Section~\ref{sec:disc} interprets each finding in detail in the light of prior work.

State-of-the-art narrative models~\citep{saxena2024select,papalampidi2020} typically rely on opaque, implicit embeddings, a black-box approach that overlooks the explicit psycholinguistic cues such as surprisal and discourse structure that trigger human attention. Our contribution is that we extend prior work by building a system for identifying specific, interpretable linguistic signals that differentiate salient scenes from background exposition. In the process, we establish a rich feature set describing salience in screenplays.
\section{Related Work and Feature Descriptions}\label{sect:features}

\subsection{Operationalising Salience}

\paragraph{Salience versus narrative importance}
We operationalise salience as alignment with the human plot summary, and treat
that alignment as a trace of attention. Attentional priority is not purely
stimulus-driven but also reflects selection history and goal relevance
\citep{awh2012}; arousing content is preferentially consolidated in memory
\citep{mather2011arousal}; summarisation is itself the extraction of a text's
macro-relevant content \citep{kintsch1978toward}; and graded-salience accounts
tie salience to accessibility \citep{giora1997understanding}. Summary alignment
\citep{saxena2024select} and turning points \citep{papalampidi2019movie} are also
the standard operationalisations here, keeping our results comparable.

\paragraph{Surprisal and the importance labels}
How prediction relates to meaning and event structure is open, and we test it
rather than assume it. Recent work weights lexical surprisal by semantic content
to fit brain responses \citep{russo2020semantics}, asks whether surprisal and
semantic relatedness contribute separably to reading
\citep{salicchi2023surprisal}, and shows that Bayesian surprise tracks where
listeners place event boundaries \citep{kumar2023bayesian}; these are event transitions
that a plot summary is likely to record, tying summary-based labels to narrative
event structure. Building on this and on accounts relating salience to
predictability \citep{salience-zarcone2016}, \S\ref{sec:corpus_study} asks
whether scene-aggregate surprisal survives length and position controls, and
\S\ref{sec:analysis} examines what remains at the token level.

\subsection{Feature Descriptions}

\citet{jaeger2010redundancy} validated the UID hypothesis on {\em that}-complementizer choice in English conversational speech using a multi-factorial analysis~\citep{baayen2008}. Their statistical model included controls reflecting competing hypotheses. This design served as an inspiration for our own features representing scene salience described below and motivated from prior work in computational linguistics, cognitive science, and narratology (see Appendix~\ref{app:feature_specs} for details of feature implementation).

\paragraph{Scene Length}
We extract raw token count, $\log(\text{num\_words})$ (used as both feature and length-control covariate), and sentence count.

\paragraph{Surprisal and Cognitive Processing}
We estimate surprisal across three structural levels following Surprisal Theory \citep{hale2001probabilistic,levy2008expectation}: 1.\ Characters \citep{oh2021}, using GPT-2 probabilities for orthographic complexity. 2.\ Words, via KenLM \citep{heafield2011kenlm} and BERT pseudo-log-likelihoods \citep{salazar2020masked}. 3.\ Chunks, using the \textsc{PsychFormers} toolkit. We also compute per-feature variance to operationalise the Uniform Information Density Hypothesis \citep{jaeger2010redundancy}.

\paragraph{Morphosyntactic Structure}
Syntax, tense, and aspect mark narrative transitions and modulate salience \citep{moens1988temporal, gibson1998linguistic}. Building on \citet{papalampidi2020} and Reichenbach's temporal framework \citep{reichenbach1947elements}, we extract three groups: 1.\ Syntactic complexity: argument-adjunct ratios, tree depth, head-dependent distance, and dependency bigrams. 2.\ Grammatical style: part-of-speech ratios (e.g., nouns, verbs). 3.\ Temporal anchoring: past/present-tense verb counts, temporal adverbs, and tense dominance ratios.

\vspace{-2pt}

\paragraph{Lexical Specificity and Stylistic Register}
Lexical diversity and readability distinguish plot-critical events from transitions \citep{mccarthy2010mtld, flesch1948new, gorinski2015movie}. We characterize this through three groups: 1.\ Lexical diversity: type-token ratio (TTR) and character-level orthographic variation indices. 2.\ Pacing and style: Flesch-Kincaid readability, sentence length variance, and punctuation cues. 3.\ Screenplay-specific vocabulary: density of action, sensory, and speech verbs \citep{elson2010extracting}.

\vspace{-2pt}

\paragraph{Semantic Grounding and Register}
The Graded Salience Hypothesis~\citep{giora1997understanding} proposes that more salient meanings are accessed faster, suggesting that conceptually intense conflict scenes carry higher narrative importance. Our data shows salient scenes contain significantly more abstract language (Appendix~\ref{app:feature_stats}). We jointly model word concreteness \citep{brysbaert2014concreteness} and sentiment polarity, which distinguish grounded action from abstract dialogue and proxy perceptual anchoring of events.

\vspace{-2pt}

\paragraph{Discourse-Pragmatic Structure}
Rhetorical Structure Theory (RST) locates core narrative content in discourse nuclei and secondary information in satellites \citep{mann1988rhetorical}; structural position and plot connectivity predict importance \citep{zeldes2026, papalampidi2020, saxena2024select, tsipidi2024surprise}. We model these dynamics via three indicators: 1.\ Rhetorical complexity: RST tree depth and path length from a discourse parser \citep{chistova-2025-bridging}. 2.\ Plot connectivity: TextRank centrality \citep{Mihalcea2004} on semantic graphs to identify globally central scenes. 3.\ Rhetorical relations: counts of causal and contrastive connectives indexing plot transitions \citep{elson2010extracting}.

\vspace{-2pt}

\paragraph{Dialogue and Deixis}
Plot-driving scenes balance action descriptions with character dialogue \citep{mckee1997story, elson2010extracting}, and sustained referential focus on specific participants increases discourse prominence \citep{ariel1990accessing, chambers2008unsupervised}. Leveraging the notion that ``{\it what is said is salient}''~\citep{leon2026}, we capture these dynamics via two measures: 1.\ Dialogue density: the ratio of speech to action description. 2.\ Referential focus: the distribution of first-, second-, and third-person pronouns, distinguishing immediate engagement from detached observation.

\vspace{-2pt}

\paragraph{Emotional Affect}
Emotional arousal amplifies salience \citep{mather2011arousal, wang2013}, and volatile scenes often mark turning points where fear, anticipation and surprise cluster~\citep{reagan2016emotional}. We capture this through three features: 1.\ Sentiment polarity: mean and variance, reflecting emotional volatility. 2.\ Categorical emotion: emotional texture via the NRC word-emotion lexicon \citep{mohammad2013crowdsourcing}. 3.\ Emotional trajectories: polarity shift slopes and magnitudes, separating dynamic beats from static atmosphere.

\vspace{-2pt}

\paragraph{Narrative Structure}
Salient scenes cluster at structural positions within three-act frameworks \citep{papalampidi2020}, with suspense a complementary signal \citep{wilmot2020suspense, berlot2026}; importance also stems from character convergence, introductions and lexical novelty signalling plot shifts \citep{bamman2013learning, gorinski2015movie}. We capture global structure through four components: 1.\ Screenplay structure: normalized position and three-phase segmentation. 2.\ Plot shifts: lexical overlap with neighboring scenes, vocabulary novelty and dialogue turnover. 3.\ Thematic callbacks: lexical similarity to the film's opening and closing scenes. 4.\ Character progression: turnover, introductions and reintroductions via named-entity recognition \citep{bamman2013learning, gorinski2015movie, elson2010extracting}.
\vspace{-5pt}
\section{Dataset}
We use the \textsc{MENSA} corpus of screenplays~\cite{saxena2024select}, a large-scale benchmark containing 924 movies (approx.\ 133k scenes). The dataset uses a realistic ``noisy supervision'' regime to identify salient scenes for each movie: the training split (824 movies) relies on automatic \textit{silver} labels, while the validation and test splits (50 movies each) use high-quality \textit{gold} human annotations. The silver labels are the automatically aligned training labels distributed with the corpus, obtained by aligning each scene to the movie's Wikipedia plot summary rather than by human annotation \citep{saxena2024select}; a scene is labeled \textit{salient} if such an alignment is found. The gold labels apply the same criterion, but the alignment is verified by human annotators. The data is characterized by severe class imbalance (only $\approx$16\% salient scenes) and significant distribution shifts between the silver training and gold evaluation sets. Additionally, salient scenes are significantly longer than non-salient ones (mean 355.7 vs.\ 132.8 words), introducing a strong verbosity bias. We provide detailed split statistics, distribution plots, and bias analysis in Appendix~\ref{app:dataset_details}.
\section{Corpus-Level Regression Analysis}\label{sec:corpus_study}

We ask which scene-level features carry independent salience signal once collinearity, positional structure, and scene length are controlled. We fit a class-balanced (within-movie) logistic regression on the gold-labelled MENSA validation split (the appropriate site for hypothesis testing, unlike the silver-aligned training split; \citealp{saxena2024select}) with the test split for replication, and apply post-LASSO inference \citep{tibshirani1996regression,belloni2013least}: $\ell_1$ regularisation selects a sparse predictor set, an unregularised refit yields standardised coefficients and Wald statistics, and $\log(\text{num\_words})$ is force-included as a length control. Full setup in Appendix~\ref{app:corpus_setup}.

\paragraph{Positional features dominate.} Three positional or structural features dominate the refit by a substantial margin, with $|z|$ values ($5.2$--$7.9$) well separated from the next tier ($|z| < 3.6$): a scene's normalised within-movie index, its proximity to the narrative edges, and its centrality in the character co-occurrence graph (Table~\ref{tab:cs_post_lasso}), aligning with the turning-point framework of \citet{papalampidi2019movie,papalampidi2020}. No linguistic feature reaches a comparable magnitude.

\begin{table}[t]
\centering
\footnotesize
\setlength{\tabcolsep}{4pt}
\begin{tabular}{lrrr}
\toprule
\textbf{Feature} & \textbf{Coef.} & \textbf{$z$} & \textbf{$p$} \\
\midrule
\multicolumn{4}{l}{\textit{Top features by $|z|$:}} \\
scene\_index\_norm     & $+0.36$ & $+7.88$ & $3.3{\times}10^{-15}$ \\
pos\_edge\_proximity   & $-0.18$ & $-5.21$ & $1.9{\times}10^{-7}$ \\
centrality\_score      & $+0.23$ & $+5.17$ & $2.4{\times}10^{-7}$ \\
\midrule
\multicolumn{4}{l}{\textit{Length control:}} \\
$\log(\text{num\_words})$ & $+0.48$ & $+2.33$ & $0.020$ \\
\bottomrule
\end{tabular}
\caption{Post-LASSO Wald inference on $\ell_1$-selected predictors plus the forced length control. Fit on the MENSA \textbf{validation split} (gold human annotations; $8{,}052$ scenes from $50$ movies, of which $2{,}677$ ($33.25\%$) are salient). Full top-100 ranking in Appendix~\ref{app:corpus_setup}.}
\label{tab:cs_post_lasso}
\end{table}

\paragraph{No scene-aggregate UID effect.} The pool included 29 BERT scene-aggregate UID statistics (variance, upper-tail surprisal, mean, IQR, kurtosis, and related summaries). The $\ell_1$ step zeroed every one on both splits, so their univariate effects ($d$ in the $-0.4$ to $-0.5$ range) reflect signal fully absorbed by scene length and within-movie position. This aligns with \citet{juzek2024signal}, who find that neither $\Delta$-UID nor $\mu$-UID outperforms random baselines on syntactic choice across 936 English sentence pairs. The token-level analysis (\S\ref{sec:disc}) qualifies the negative result: a UID-like pattern persists, but only as asymmetric suppression of upper-tail per-token surprisal. 

\paragraph{RST and word-order features.} They surface in the top-100 ranking (Appendix~\ref{app:corpus_setup}) with effect sizes well below the dominant positional features, consistent with the precision-filter role observed in \S\ref{sec:rst_utility}. These results motivate combining structural information directly with BERT scene representations, which the LGSM (\S\ref{sect:methods}) instantiates through adaptive gated fusion.
\section{The Linguistically-Grounded Salience Model (LGSM)}\label{sect:methods}

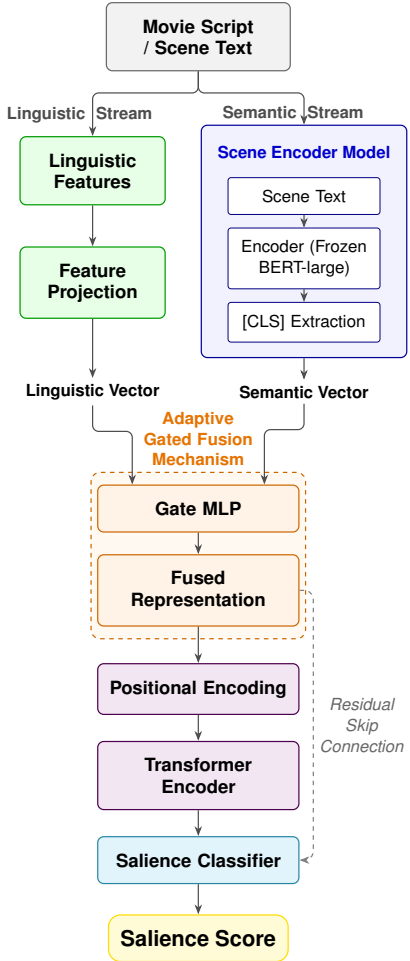
\begin{figure}[t]
\centering
\resizebox{0.70\columnwidth}{!}{%
\hyphenpenalty=10000\exhyphenpenalty=10000\relax
\begin{tikzpicture}[
  font=\sffamily\large, 
  >=Stealth,
  every node/.style={align=center},
  io/.style={
    draw=gray!80!black, thick, rounded corners=4pt,
    fill=gray!10, text=black,
    inner xsep=8pt, inner ysep=11pt,
    text width=3.7cm, minimum height=0.8cm
  },
  ling/.style={
    draw=green!60!black, thick, rounded corners=4pt,
    fill=green!10, text=black,
    inner xsep=8pt, inner ysep=11pt,
    text width=2.8cm, minimum height=0.8cm
  },
  sem/.style={
    draw=blue!60!black, thick, rounded corners=4pt,
    fill=blue!5, text=black,
    inner xsep=7pt, inner ysep=10pt,
    minimum width=3.8cm
  },
  zoomnode/.style={
    draw=blue!50!black, fill=white, rounded corners=2pt,
    font=\sffamily\normalsize, inner xsep=6pt, inner ysep=8pt, 
    text width=3.1cm
  },
  fuse/.style={
    draw=orange!80!black, thick, rounded corners=4pt,
    fill=orange!10, text=black,
    inner xsep=8pt, inner ysep=11pt,
    text width=4.1cm, minimum height=0.85cm
  },
  seq/.style={
    draw=violet!60!black, thick, rounded corners=4pt,
    fill=violet!10, text=black,
    inner xsep=8pt, inner ysep=11pt,
    text width=4.1cm, minimum height=0.85cm
  },
  clf/.style={
    draw=cyan!60!black, thick, rounded corners=4pt,
    fill=cyan!10, text=black,
    inner xsep=8pt, inner ysep=11pt,
    text width=4.1cm, minimum height=0.85cm
  },
  outnode/.style={
    draw=yellow!80!orange, thick, rounded corners=8pt,
    fill=yellow!20, text=black,
    inner xsep=8pt, inner ysep=10pt,
    font=\sffamily\Large\bfseries 
  },
  lbl/.style={font=\sffamily\normalsize\bfseries, text=black!70},
  veclbl/.style={font=\sffamily\normalsize\bfseries, text=black},
  dashedbox/.style={
    draw=orange!80!black, thick, dashed, rounded corners=6pt,
    inner xsep=4pt, inner ysep=8pt
  },
  arr/.style={->, thick, black!70},
  skip/.style={->, thick, black!50, dashed},
]

\pgfdeclarelayer{background}
\pgfsetlayers{background,main}

\node[io] (input) at (0,0) {\textbf{Movie Script / Scene Text}};

\node[ling, below=1.5cm of input, xshift=-2.45cm] (lf) 
  {\textbf{Linguistic Features}};
\node[ling, below=0.95cm of lf] (fp) 
  {\textbf{Feature Projection}}; 

\node[font=\sffamily\normalsize\bfseries, text=blue!80!black, below=1.6cm of input, xshift=2.45cm] (sem_title) {Scene Encoder Model};
\node[zoomnode, below=0.3cm of sem_title] (z_txt) {Scene Text};
\node[zoomnode, below=0.3cm of z_txt] (z_enc) {Encoder (Frozen BERT-large)};
\node[zoomnode, below=0.3cm of z_enc] (z_cls) {[CLS] Extraction};

\draw[arr] (z_txt) -- (z_enc);
\draw[arr] (z_enc) -- (z_cls);

\begin{scope}[on background layer]
  \node[sem, fit=(sem_title) (z_txt) (z_cls)] (se) {};
\end{scope}

\node[lbl] at ([xshift=-0.3cm, yshift=0.5cm]lf.north) {Linguistic\hspace{0.25cm}Stream};
\node[lbl] at ([xshift=-0.25cm, yshift=0.3cm]se.north) {Semantic\hspace{0.25cm}Stream};

\node[veclbl, below=0.5cm of se] (hsem) {Semantic Vector};
\node[veclbl] (hling) at (fp.south |- hsem.center) {Linguistic Vector};

\draw[arr] (fp.south) -- (hling.north);
\draw[arr] (se.south) -- (hsem.north);

\coordinate (mid) at ($(hling)!0.5!(hsem)$);

\node[fuse, below=2.15cm of mid] (gate) 
  {\textbf{Gate MLP}}; 
\node[fuse, below=0.5cm of gate] (fused) 
  {\textbf{Fused Representation}}; 

\begin{scope}[on background layer]
  \node[dashedbox, fill=orange!5, fit=(gate)(fused)] (fusionbox) {};
\end{scope}

\node[lbl, text=orange!90!black, fill=white, inner sep=1pt, text width=3.4cm, anchor=south, yshift=3pt] at (fusionbox.north) {Adaptive Gated Fusion Mechanism};

\node[seq, below=0.55cm of fusionbox] (pe) 
  {\textbf{Positional Encoding}}; 

\node[seq, below=0.6cm of pe] (tx) 
  {\textbf{Transformer Encoder}}; 

\node[clf, below=0.6cm of tx] (class) 
  {\textbf{Salience Classifier}}; 

\node[outnode, below=0.7cm of class] (output) {Salience Score}; 

\draw[arr, rounded corners=4pt] (input.south) -- ++(0,-0.5) -| (lf.north);
\draw[arr, rounded corners=4pt] (input.south) -- ++(0,-0.5) -| (se.north);

\draw[arr] (lf.south) -- (fp.north);

\draw[arr, rounded corners=4pt] (hling.south) -- ++(0,-0.6) -| (gate.160);
\draw[arr, rounded corners=4pt] (hsem.south) -- ++(0,-0.6) -| (gate.20);
\draw[arr] (gate.south) -- (fused.north);

\draw[arr] (fused.south) -- (pe.north);
\draw[arr] (pe.south)    -- (tx.north);
\draw[arr] (tx.south)    -- (class.north);
\draw[arr] (class.south) -- (output.north);

\draw[skip, rounded corners=6pt] 
  (fused.east) -- ++(0.32,0) |- node[pos=0.25, right, align=center, font=\sffamily\normalsize\itshape, text=black!60] {Residual \\ Skip \\ Connection} (class.east);


\end{tikzpicture}%
}
\caption{Linguistically-Grounded Salience Model (LGSM)}
\label{fig:architecture}
\end{figure}

We formulate salience detection as binary sequence labeling: predict $y_t \in \{0, 1\}$ for each scene $S_t$ in a script $\mathcal{M}$. LGSM (Figure \ref{fig:architecture}) is designed to measure, not merely exploit, the network's reliance on explicit psycholinguistic features versus implicit semantic embeddings. We keep the two separate because they answer different questions. A single encoder mixes surface, syntactic and semantic information across its layers, so no one cue's contribution can be read off directly. Supplying the features in a separate stream the model may weight or ignore makes that reliance measurable: the gate $\alpha_t$ below states, per scene, how much of the prediction rests on them. Interpretability, not accuracy, is what motivates the separation.

\vspace{-2pt}
\paragraph{Dual-Stream Scene Encoding:}
Each scene is processed through two parallel streams.
\vspace{-2pt}
\paragraph{The Semantic Stream:}
We encode scene $S_t$ with a frozen \texttt{bert-large} encoder, taking the \texttt{[CLS]} token's final hidden state: $h_{\text{sem}} = \text{BERT}(S_t)_{\text{[CLS]}} \in \mathbb{R}^{d_{\text{bert}}}$. Freezing establishes a constant semantic baseline, so any structural adaptation is driven entirely by the explicit features.

\vspace{-4pt}
\paragraph{The Linguistic Stream.}
We project the explicit feature vector $x_{\text{ling}} \in \mathbb{R}^k$ into a latent space compatible with the semantic embeddings: $h_{\text{ling}} = \text{GELU}(W_p x_{\text{ling}} + b_p) \in \mathbb{R}^{d_{\text{ling}}}$.

\paragraph{Adaptive Gated Fusion:} Concatenation screens feature importance, so we mediate between the streams with a scalar gate $\alpha_t \in (0, 1)$ quantifying reliance on the linguistic signal:
\(\alpha_t = \sigma(W_g [h_{sem}; h_{ling}] + b_g)\)
The fused embedding is a convex combination of the streams:
\(h_{fused} = \alpha_t \cdot (W_l h_{ling}) + (1 - \alpha_t) \cdot h_{sem}\). Analyzing $\alpha_t$ allows us to transparently observe when the network prioritizes structural cues over surface semantics.

\paragraph{Contextualization and Optimization:} The sequence of fused embeddings $H = \{h_{\text{fused}}^{(1)}, \dots, h_{\text{fused}}^{(N)}\}$ passes through a 2-layer Transformer Encoder, modelling global plot dependencies, with unmasked self-attention and sinusoidal positional embeddings $P$. Given the extreme class imbalance, we optimize with Focal Loss \cite{Lin2017} ($\gamma=3.0$, $\alpha=0.25$). The linguistic stream relies on computationally intensive upstream extractors (GPT-2, KenLM, neural RST parsers), but we incur no runtime overhead: extraction is strictly decoupled from sequence modeling, and all explicit features and semantic embeddings are pre-computed and cached offline (Appendix \ref{app:computational_complexity}).

Full training and implementation details, including optimiser, schedule and feature preprocessing, are given in Appendix~\ref{app:implementation}. We turn next to LGSM's classification performance against the baselines (\S\ref{sec:results}) and to the analyses that probe what the architecture relies on (\S\ref{sec:analysis}).
\section{Results}
\label{sec:results}

We compare LGSM against unsupervised, heuristic, and supervised baselines, report headline classification numbers, and validate the recovered scenes through downstream summarisation. Narrative-arc stability and gating behaviour are analysed in \S\ref{sec:arc_gating}.

\paragraph{Baselines.} \textit{Select and Summarize} \citep{saxena2024select} is a purely semantic neural baseline using a pre-trained RoBERTa encoder followed by a temporal-aware Transformer; it serves as the direct control to measure the value of adding explicit linguistic priors. \textit{Logistic Regression} is a linear classifier trained on the scalar linguistic features alone, validating their independent predictive power. \textit{Weighted Ensemble} is a soft-voting ensemble of 13 classifiers (LR, Random Forest, and SVMs with Linear/RBF/Poly/Sigmoid kernels) trained on the linguistic features to test non-neural limits. We additionally include zero-shot generative baselines from Qwen and Gemini 2.5 Flash. Implementation details for all baselines are given in Appendix~\ref{app:baselines}.

\begin{table}[t]
\centering
\footnotesize
\renewcommand{\arraystretch}{1.1}
\setlength{\tabcolsep}{2pt}
\begin{tabular}{@{}>{\raggedright\arraybackslash}p{2.8cm}|ccc|c@{}}
\hline
\multicolumn{1}{c|}{\multirow{2}{*}{\textbf{Model}}} & \multicolumn{3}{c|}{\textbf{Salient Class (Binary)}} & \textbf{Overall} \\
\cline{2-4} \cline{5-5}
 & \textbf{Prec.} & \textbf{Recall} & \textbf{F1} & \textbf{Macro F1} \\
\hline
\vspace{1pt}
\textit{Baselines} & & & & \\
Majority Class & 0.00 & 0.00 & 0.00 & 38.70 \\
Weighted Ensemble & 56.67 & 56.10 & 56.38 & 67.55 \\
Logistic Regression & 54.58 & 59.45 & 56.91 & 67.14 \\
Select and Summarize & 58.93 & 51.56 & 55.00 & \textbf{67.70} \\
Qwen & 42.60 & 46.40 & 44.40 & 57.41 \\
Gemini 2.5 Flash (Zero-Shot Prompt) & \textbf{63.20} & 46.40 & 53.51 & 67.40 \\
\hline
\vspace{1pt}
\textit{Loss control (BERT only)} & & & & \\
\quad Weighted BCE & 55.70 & 53.05 & 54.34 & 66.34 \\
\quad Focal Loss & 54.56 & 53.65 & 54.10 & 65.91 \\
\hline
\vspace{1pt}
\textit{Proposed Method (LGSM)} & & & & \\
Weighted BCE & 51.72 & 64.29 & 57.32 & 66.91 \\
w/o semantics & 52.27 & 64.75 & 57.86 & 66.00 \\
w/o surprisal & 53.49 & 64.79 & 58.60 & 66.85 \\
w/o narrative-structure & 52.97 & 64.88 & 58.32 & 65.91 \\
w/o RST & 47.48 & 72.69 & 57.45 & 63.50 \\
\textbf{Focal Loss} & 51.91 & \textbf{68.34} & \textbf{59.01} & 67.08 \\
\hline
\end{tabular}
\caption{Salience identification on the MENSA test set (8156 scenes, 50 movies).}
\label{tab:classification_results}
\end{table}

\paragraph{Classification results.} The integration of explicit linguistic features increases recall on the salient class from $51.56\%$ (Select and Summarize semantic baseline) to $\mathbf{68.34\%}$ for LGSM with Focal Loss (Table~\ref{tab:classification_results}), an absolute gain of $+16.78$ points. Macro F1 is approximately matched ($67.08$ vs.\ $67.70$). The trade-off is precision: LGSM with Focal Loss achieves $51.91\%$ precision against $58.93\%$ for the semantic baseline. We accept this trade-off because in narrative extraction a missed salient scene (false negative) breaks the story chain, whereas an extra non-salient scene (false positive) adds context that is recoverable downstream. Zero-shot generative baselines like Gemini 2.5 Flash achieve high precision ($63.20\%$) but recall only $46.40\%$, acting as strictly conservative filters. Ablation rows in Table~\ref{tab:classification_results} show that removing the surprisal, narrative-structure, or semantics feature groups reduces recall to roughly $64.8\%$; removing RST lifts recall to $72.69\%$ at a substantial precision cost ($47.48\%$). Focal Loss outperforms Weighted BCE on recall ($68.34\%$ vs.\ $64.29\%$) by forcing the model to focus on rare hard examples.

\paragraph{Downstream summarisation.} To validate that LGSM's recovered scenes carry information rather than noise, we fine-tune an LED-Large summariser \citep{Beltagy2020} on scenes extracted by LGSM vs.\ the semantic baseline. LGSM extractions improve ROUGE-1 by $\mathbf{+3.67}$ points and ROUGE-L by $\mathbf{+1.51}$ points over the semantic baseline, with a stable BERTScore ($83.17$ vs.\ $83.29$); the full table is in Appendix~\ref{app:summarization_results}. A detailed error decomposition (TP/FN/FP/TN breakdown, false-positive signature, false-negative outlier taxonomy) is reported in Appendix~\ref{app:error_analysis} and supports the interpretation that LGSM's additional ``false positives'' are largely setup or tension scenes too granular for the Wikipedia summary used as silver supervision rather than noise.

\paragraph{The gain is not a coverage artifact.} Because LGSM marks more scenes salient, a larger input could raise ROUGE on volume alone. Of the extra scenes it routes to the summariser, $451$ are true positives the baseline misses and $736$ are false positives ($1{,}837$ vs.\ $1{,}386$ and $1{,}702$ vs.\ $966$; Table~\ref{tab:error_comparison}), so the input has higher salient-scene recall, not merely more scenes. LED is abstractive with a fixed maximum decoding length, so passing it roughly $50\%$ more scenes ($3{,}539$ vs.\ $2{,}352$) raises the compression ratio at constant output length: a ROUGE gain reflects content composition rather than input volume. The per-metric pattern agrees: a volume effect would lift every $n$-gram order together, whereas ROUGE-2 dips $0.27$ while ROUGE-1 and ROUGE-L rise.
\section{Analysis}\label{sec:analysis}

This section examines what distinguishes salient from non-salient scenes through ablations and SHAP attribution with split-level and token-level follow-ups.



\subsection{Conditions for Feature Utility: RST}\label{sec:rst_utility}
\label{sec:feature_utility}
Ablation of the discourse-pragmatic (RST) group (Table~\ref{tab:classification_results}) shows it functions as a precision filter rather than a recall booster: removing it increases recall (64.29\%$\rightarrow$72.69\%) but substantially degrades precision (51.72\%$\rightarrow$47.48\%), indicating that RST features suppress false positives on the larger non-salient class. This is consistent with the gating diagnostics for Movie 17 (Appendix~\ref{app:gating_diagnostics}), where higher linguistic-stream weighting tracks non-salient predictions; most other movies show near-zero gate-prediction correlation, so we treat this as suggestive rather than general confirmation of the mechanism.

\subsection{SHAP-Based Feature Analysis}\label{sec:shap_analysis}

SHAP attribution points to a small set of interpretable features (Figure~\ref{fig:shap_correlation}; the attribution procedure is detailed in Appendix~\ref{app:shap_procedure}). On validation, salient scenes have lower BERT surprisal aggregates, more than twice as many central characters ($5.66$ vs.\ $2.48$), higher third-person pronoun counts ($12.83$ vs.\ $4.96$), and lower readability. Held-out test checks reveal which signals are stable: per-scene BERT variance is largely a length effect, while the stable within-scene surprisal signal sits in the upper tail. Character and style signals are simpler and match prior work on protagonist-centered narrative importance. Full SHAP patterns and validation feature statistics are in Appendices~\ref{app:shap_contributions} and~\ref{app:feature_stats}.

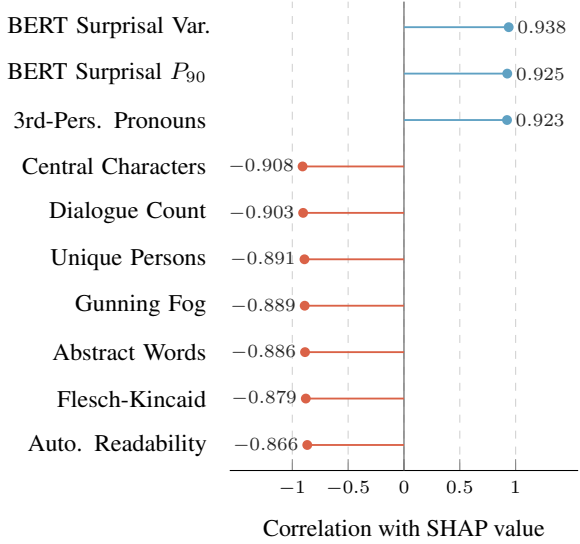
\begin{figure}[t]
\centering
\hyphenpenalty=10000\exhyphenpenalty=10000\relax
\begin{tikzpicture}
  \begin{axis}[
    xlabel={Correlation with SHAP value},
    xlabel style={font=\footnotesize, yshift=-0.4ex},
    width=\dimexpr\columnwidth-88pt\relax, height=6.2cm,
    scale only axis,
    xmin=-1.56, xmax=1.56,
    ymin=0.45, ymax=10.55,
    xtick={-1,-0.5,0,0.5,1},
    xticklabel style={font=\scriptsize},
    ytick={1,...,10},
    yticklabels={
        {Auto. Readability},
        {Flesch-Kincaid},
        {Abstract Words},
        {Gunning Fog},
        {Unique Persons},
        {Dialogue Count},
        {Central Characters},
        {3rd-Pers. Pronouns},
        {BERT Surprisal $P_{90}$},
        {BERT Surprisal Var.}
    },
    yticklabel style={font=\footnotesize, align=right, text width=2.63cm, xshift=-5pt},
    ytick style={draw=none},
    axis x line*=bottom,
    axis y line*=left,
    y axis line style={draw=none},
    xmajorgrids=true,
    grid style={thin, dashed, black!18},
    clip=false,
    nodes near coords,
    point meta=x,
    nodes near coords style={
        /pgf/number format/fixed,
        /pgf/number format/precision=3,
        /pgf/number format/zerofill,
        font=\scriptsize, text=black!85, inner sep=1pt
    },
  ]

  \draw[black!60, line width=0.45pt] (axis cs:0,0.45) -- (axis cs:0,10.55);

  \addplot[xcomb, color=barred, line width=0.7pt,
           mark=*, mark size=1.5pt, mark options={fill=barred, draw=barred},
           nodes near coords align={left},
           every node near coord/.append style={xshift=-2pt}]
    coordinates {
    (-0.908, 7) (-0.903, 6) (-0.891, 5) (-0.889, 4)
    (-0.886, 3) (-0.879, 2) (-0.866, 1)
  };

  \addplot[xcomb, color=barblue, line width=0.7pt,
           mark=*, mark size=1.5pt, mark options={fill=barblue, draw=barblue},
           nodes near coords align={right},
           every node near coord/.append style={xshift=2pt}]
    coordinates {
    (0.938, 10) (0.925, 9) (0.923, 8)
  };

  \end{axis}
\end{tikzpicture}%
\caption{Correlation between feature values and SHAP attributions for the top ten features. BERT surprisal variance has the strongest correlation ($r = 0.938$), followed by upper-tail surprisal ($r = 0.925$) and third-person pronoun count ($r = 0.923$).}
\label{fig:shap_correlation}
\end{figure}

\begin{figure}[tb]
\centering
\begin{tikzpicture}
  \begin{axis}[
    width=\dimexpr\columnwidth-43pt\relax, height=4.4cm,
    scale only axis,
    xlabel={Word surprisal (nats)},
    ylabel={Cumulative probability},
    xlabel style={font=\footnotesize, yshift=1pt},
    ylabel style={font=\footnotesize, yshift=-3pt},
    xmin=0, xmax=14, ymin=0, ymax=1.02,
    xtick={0,2,4,6,8,10,12,14},
    ytick={0,0.2,0.4,0.6,0.8,1.0},
    xticklabel style={font=\scriptsize},
    yticklabel style={font=\scriptsize, /pgf/number format/fixed,
                      /pgf/number format/precision=1,
                      /pgf/number format/fixed zerofill},
    tick align=outside, tick pos=left,
    legend style={font=\scriptsize, at={(0.97,0.05)}, anchor=south east,
                  draw=none, fill=none, row sep=-2pt, inner sep=1pt},
    legend cell align=left,
  ]

  \addplot[draw=none, fill=black!50, fill opacity=0.09, forget plot]
    coordinates {(2.3354,0) (14,0) (14,1.02) (2.3354,1.02)} \closedcycle;

  \addplot[barblue, line width=0.5pt, densely dotted, forget plot]
    coordinates {(5.3475,0) (5.3475,1.02)};
  \addplot[barred, line width=0.5pt, densely dotted, forget plot]
    coordinates {(5.9053,0) (5.9053,1.02)};

  \addplot[barred, line width=1.1pt, mark=none] coordinates {
      (-0.000,0.0026) (0.014,0.2927) (0.028,0.3413) (0.056,0.3926)
      (0.084,0.4240) (0.112,0.4467) (0.154,0.4722) (0.196,0.4924)
      (0.252,0.5137) (0.308,0.5313) (0.364,0.5462) (0.448,0.5650)
      (0.519,0.5786) (0.631,0.5976) (0.715,0.6099) (0.841,0.6264)
      (0.995,0.6442) (1.149,0.6599) (1.303,0.6744) (1.499,0.6912)
      (1.850,0.7179) (2.060,0.7324) (2.312,0.7486) (2.537,0.7621)
      (3.027,0.7892) (3.742,0.8240) (3.882,0.8301) (4.162,0.8418)
      (4.779,0.8651) (5.157,0.8779) (5.886,0.8996) (6.138,0.9062)
      (6.615,0.9182) (6.909,0.9248) (7.792,0.9421) (8.871,0.9584)
      (9.558,0.9669) (9.964,0.9710) (10.370,0.9749) (11.253,0.9823)
      (11.688,0.9853) (12.164,0.9881) (12.753,0.9908) (14.000,0.9948)
  };
  \addlegendentry{Non-salient}

  \addplot[barblue, line width=1.1pt, mark=none] coordinates {
      (-0.000,0.0026) (0.014,0.3165) (0.028,0.3673) (0.056,0.4201)
      (0.084,0.4522) (0.126,0.4847) (0.168,0.5085) (0.224,0.5326)
      (0.280,0.5518) (0.336,0.5674) (0.392,0.5810) (0.476,0.5987)
      (0.547,0.6113) (0.631,0.6249) (0.729,0.6390) (0.869,0.6567)
      (1.037,0.6752) (1.163,0.6877) (1.303,0.7004) (1.499,0.7166)
      (1.752,0.7354) (1.948,0.7490) (2.214,0.7660) (2.719,0.7951)
      (3.307,0.8245) (3.966,0.8530) (4.246,0.8638) (4.639,0.8778)
      (4.877,0.8858) (5.227,0.8966) (5.409,0.9018) (5.872,0.9142)
      (6.306,0.9247) (6.993,0.9390) (7.512,0.9482) (8.212,0.9585)
      (8.857,0.9665) (9.502,0.9731) (10.216,0.9792) (11.155,0.9856)
      (12.444,0.9917) (14.000,0.9959)
  };
  \addlegendentry{Salient}

  \end{axis}
\end{tikzpicture}
\caption{CDFs of word-level BERT surprisal (validation set). Curves track
         closely below the median; salient rises faster above it, with a
         $0.56$-nat horizontal gap at $P_{90}$.}
\label{fig:token_level_cdf}
\end{figure}
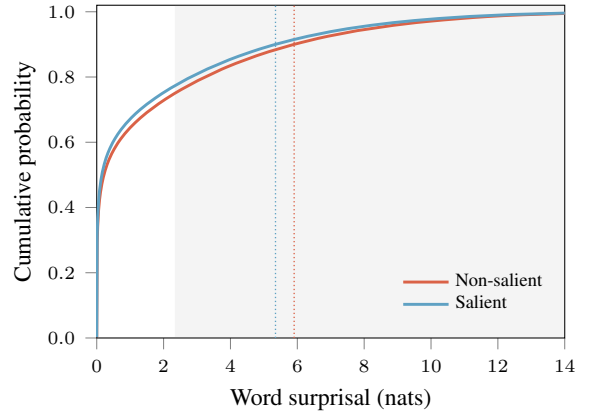

\paragraph{Surprisal signals and what they track.}
BERT variance has the highest SHAP attribution: on validation salient scenes show lower variance ($d = -0.41$; Table~\ref{tab:bert_surprisal_stats}), suggesting smoother per-token surprisal contours. The test split complicates this: variance more than halves ($d = -0.41 \to -0.16$) while upper-tail surprisal stays stable ($d = -0.45 \to -0.40$). A drop that large indicates the feature tracks length, not salience: longer scenes tend to have lower per-token surprisal variance in this corpus, and salient MENSA scenes are longer than non-salient ones on average. A quantile is bounded by the empirical tail rather than averaged over the scene, so the upper-tail aggregate is less length-sensitive; Appendix~\ref{app:bert_surprisal} reports the additional validation of these surprisal features. Token-level analysis (Table~\ref{tab:token_level_uid}, Figure~\ref{fig:token_level_cdf}) confirms this: the two distributions are nearly identical in the bulk (medians within $0.07$ nats, pooled Cohen's $d$ near zero, $97\%$ histogram overlap), with the difference concentrated in the upper tail, where salient tokens are $0.5$--$0.65$ nats lower at $P_{90}$ and $P_{95}$ (Appendix~\ref{app:token_level_distributions}). The small mean shift between classes ($\Delta\text{mean} \approx -0.18$ nats) is driven almost entirely by this upper-tail suppression rather than a broad shift to easier vocabulary: salient scenes specifically avoid the rarest, hardest-to-predict tokens while their bulk vocabulary is essentially indistinguishable from non-salient material (see Section~\ref{sec:disc} for plausible explanations). 

\begin{table}[ht!]
\centering
\footnotesize
\setlength{\tabcolsep}{3pt}
\begin{tabular}{l l r r r r}
\toprule
\textbf{Feature} & \textbf{Split} & \textbf{Sal.} & \textbf{Non-sal.} & \textbf{$d$} & \textbf{$p$} \\
\midrule
Variance     & val  & $8.34$  & $10.31$ & $-0.41$ & $<10^{-90}$ \\
Variance     & test & $10.30$ & $11.69$ & $-0.16$ & $<10^{-57}$ \\
\addlinespace
Upper-tail   & val  & $5.71$  & $6.54$  & $-0.45$ & $<10^{-90}$ \\
Upper-tail   & test & $5.76$  & $6.49$  & $-0.40$ & $<10^{-60}$ \\
\bottomrule
\end{tabular}
\caption{BERT per-scene variance and upper-tail (90th percentile) surprisal, salient vs.\ non-salient. Variance halves from val to test; upper-tail stays stable.}
\label{tab:bert_surprisal_stats}
\end{table}
\vspace{0.4\baselineskip}

\begin{table}[ht!]
\centering
\footnotesize
\setlength{\tabcolsep}{5pt}
\begin{tabular}{l r r}
\toprule
\textbf{Quantity (salient $-$ non-salient)} & \textbf{Val} & \textbf{Test} \\
\midrule
$\Delta P_5$ (nats)    & ${\approx}0$ & ${\approx}0$ \\
$\Delta P_{25}$ (nats) & ${\approx}0$ & ${\approx}0$ \\
$\Delta$\,mean (nats)  & $-0.18$      & $-0.16$      \\
$\Delta P_{90}$ (nats) & $-0.56$      & $-0.53$      \\
$\Delta P_{95}$ (nats) & $-0.65$      & $-0.61$      \\
\midrule
Pooled Cohen's $d$     & $-0.07$      & $-0.05$      \\
Histogram overlap      & $0.97$       & $0.97$       \\
\bottomrule
\end{tabular}
\caption{Token-level BERT word-surprisal, salient vs.\ non-salient. Bulk distributions match; salient tokens are $0.5$--$0.65$ nats lower at $P_{90}/P_{95}$.}
\label{tab:token_level_uid}
\end{table}

\begin{figure}[t]
\centering
\begin{tikzpicture}
  \begin{axis}[
    width=\dimexpr\columnwidth-36pt\relax, height=5.4cm,
    scale only axis,
    xlabel={Number of Central Characters},
    ylabel={Salience Probability},
    xlabel style={font=\footnotesize, yshift=1pt},
    ylabel style={font=\footnotesize, yshift=-3pt},
    xmin=-2.05, xmax=43.06, ymin=-0.05, ymax=1.05,
    xtick={0,10,20,30,40},
    ytick={0,0.2,0.4,0.6,0.8,1.0},
    yticklabel style={font=\scriptsize, /pgf/number format/fixed,
                      /pgf/number format/precision=1,
                      /pgf/number format/fixed zerofill},
    xticklabel style={font=\scriptsize},
    grid=both, grid style={thin, dotted, black!25},
    tick align=outside, tick pos=left,
    legend style={font=\scriptsize, at={(0.03,0.97)}, anchor=north west,
                  draw=black!25, fill=white,
                  row sep=-1pt, inner sep=2pt},
    legend cell align=left,
  ]

  \addplot[scatter, only marks, mark=*,
     visualization depends on={\thisrow{ms}\as\msize},
     scatter/use mapped color={draw=black!55, fill=black!62, fill opacity=0.55},
     scatter/@pre marker code/.append style={/tikz/mark size=\msize}]
   table {central_chars.dat};
  \addlegendentry{Observed (binned)}

  \addplot[blue, line width=0.9pt, mark=none] coordinates {
    (0.000,0.2275) (1.097,0.2557) (2.331,0.2900) (3.702,0.3312)
    (5.074,0.3752) (6.856,0.4354) (10.696,0.5694) (12.478,0.6294)
    (13.850,0.6731) (15.221,0.7140) (16.455,0.7480) (17.552,0.7759)
    (18.649,0.8015) (19.746,0.8249) (20.843,0.8460) (21.940,0.8650)
    (23.311,0.8859) (24.682,0.9040) (26.054,0.9194) (27.425,0.9326)
    (29.070,0.9457) (30.716,0.9564) (32.636,0.9663) (34.830,0.9750)
    (37.298,0.9821) (40.177,0.9880) (41.000,0.9892)
  };
  \addlegendentry{Linear (pseudo-$R^2\!=\!0.0735$)}

  \addplot[red, line width=0.9pt, dashed, mark=none] coordinates {
    (0.000,0.1982) (1.508,0.2606) (4.799,0.4094) (6.582,0.4862)
    (8.090,0.5442) (9.325,0.5858) (10.559,0.6218) (11.793,0.6524)
    (13.027,0.6776) (14.261,0.6979) (15.495,0.7137) (17.141,0.7282)
    (18.786,0.7358) (20.432,0.7369) (22.077,0.7315) (23.723,0.7194)
    (25.094,0.7038) (26.328,0.6851) (27.562,0.6616) (28.796,0.6329)
    (30.030,0.5988) (31.264,0.5591) (32.499,0.5139) (34.144,0.4462)
    (37.846,0.2792) (39.492,0.2094) (40.726,0.1631) (41.000,0.1537)
  };
  \addlegendentry{Quadratic (pseudo-$R^2\!=\!0.0839$)}

  \node[anchor=south east, font=\scriptsize, align=right, inner sep=1pt,
        text=black!75] at (axis cs:42.6,-0.03)
    {Quadratic coef.\ $-0.1444$\\ (standardised $x$)\\ LR test $p=7.9\times10^{-25}$};

  \end{axis}
\end{tikzpicture}
\caption{Validation-set salience probability as a function of central
         character count. Points are binned observations, sized by bin
         count; the largest bin has a mean count of $15.7$. The quadratic
         fit gives a significant inverted-U peaking near $20$ central
         characters.}
\label{fig:central_character_nonmonotonic}
\end{figure}
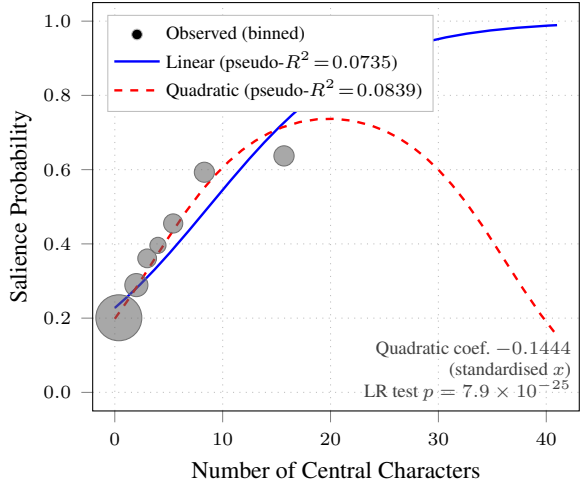

\paragraph{Character convergence, agentive focus, and stylistic register.}
Central character count is a strong predictor of salience ($d = 0.64$; Table~\ref{tab:feature_averages}), but the relationship is not monotonic: quadratic regression gives a significant inverted-U ($\beta_2 < 0$, $p < 0.001$; Figure~\ref{fig:central_character_nonmonotonic}), and the same non-monotonicity holds for unique character count (Appendix~\ref{app:nonmonotonic_character_structure}). While raw third-person singular pronoun counts strongly correlate with salience ($r = 0.923$, $d = 0.74$), their length-normalized ratio is non-significant ($r = -0.076$, $p = 0.096$), indicating the effect is driven by scene length. Readability metrics go the other way: salient scenes score lower on syntactic complexity. SHAP (Figure~\ref{fig:shap_contributions_appendix} in Appendix~\ref{app:shap_contributions}) shows that readability, dialogue density ($d = 0.69$), and abstraction ($d = 0.78$) mostly push predictions toward non-salient but occasionally the other way, indicating salience does not come from any single feature; the model picks up combinations that depend on the rest of the scene (see Section~\ref{sec:disc} for more discussion of these results).

\subsection{Narrative-arc Stability and Lost in the Middle}\label{sec:arc_gating}
As mentioned in the introduction, screenplays distribute critical events across a complex three-act structure, rendering \emph{fixed} positional heuristics ineffective; \emph{learned} positional structure remains predictive, as Section \ref{sec:corpus_study} shows. Recent studies highlight that standard models like Longformer~\citep{Beltagy2020} frequently suffer from the Lost-in-the-Middle phenomenon~\citep{Liu2024}, where performance severely degrades during central narrative segments. To test whether linguistic grounding mitigates this issue in scripts averaging 35k tokens, we evaluate recall across the normalised timeline (Acts 1-3). The semantic baseline exhibits a noticeable decrease in recall from Act 1 to Act 2A, with lower performance persisting through Act 2C, whereas LGSM maintains temporal stability throughout the entire narrative arc (Figure~\ref{fig:narrative_arc} in Appendix~\ref{app:arc_gating}). This suggests that explicitly modelling positional and structural features alongside the neural representation can mitigate the mid-document performance degradation often observed in long-form text.

We additionally analyse the gating coefficient $\alpha_t$ to evaluate how LGSM balances its two streams. Across most narratives, the fusion ratio remains remarkably consistent: Movie 28, for instance, maintains a flat temporal trend (slope $\approx 0$) tightly clustered around $\mu \approx 0.477$ (Figure~\ref{fig:gate_statistics_movie_28} in Appendix~\ref{app:gating_diagnostics}), indicating that linguistic features are continuously integrated throughout the entire film rather than isolated to specific structural acts. The gate's behaviour also adapts to the narrative: in some films $\alpha_t$ is uncorrelated with the final prediction, while in others (e.g., Movie 17) we observe a strong negative correlation ($r = -0.775$; Figure~\ref{fig:gate_statistics_movie_17} in Appendix~\ref{app:gating_diagnostics}), where higher reliance on the linguistic stream predicts non-salience. This is consistent with the model using linguistic features as a negative filter to confidently reject background scenes before shifting back to semantics for salient ones.
\section{Discussion}\label{sec:disc}

In this section, we discuss broader interpretations and implications of each of our five findings tagged in the introduction as: \textbf{(P1)}--\textbf{(N2)}.

\paragraph{(P1) Token-level upper-tail surprisal asymmetry.}
The token-level analysis (Section \ref{sec:shap_analysis}) shows that salient and non-salient scenes share nearly identical word-level surprisal distributions in the bulk while diverging in the upper tail, where salient scenes suppress the rarest tokens. This effect is consistent with \citet{salience-zarcone2016} on the asymmetric cost of chaotic prediction errors and the Structured Context Hypothesis of \citet{tsipidi2024surprise}, who present evidence that surprisal contours are shaped by hierarchical structure rather than smoothed toward a uniform mean in long-form discourse. In their critical overview of experiments testing the UID hypothesis in the literature, \citet{juzek2024signal} concludes that UID is largely effective for linguistic phenomena involving a single-word choice point (e.g., {\em that}-complementizer choice in Jaeger's work) rather than for entire scenes as in our study, a point we return to in (N1) below when discussing scene-aggregate UID summaries.



\paragraph{(P2) A recurring linguistic signature.}
A signature appears consistently across SHAP attribution and corpus-level statistics. Salience rises with central character count then falls once there are too many, matching \citet{bamman2013learning} and \citet{gorinski2015movie} on narrative importance arising from interactions among a limited set of prominent characters; the inverted-U is consistent with a ceiling on tracked-character capacity. Salient scenes are characterized by third-person pronominal reference on a small set of agents, aligning with protagonist-centered models \citep{chambers2008unsupervised, bamman2013learning}. Such scenes favour syntactically simpler constructions \citep{gibson1998linguistic}, and counterintuitively carry more abstract language, consistent with pivotal moments engaging motivation and conflict rather than concrete description \citep{mckee1997story}. Together these signals describe salient scenes as moments where a stable agent set, accessible syntax, and abstract narrative stakes converge: the configuration a comprehender can most readily integrate into a coherent situation model. In long-form fiction, salience sits where structural prominence and linguistic accessibility meet, not where vocabulary is unusual.

\paragraph{What frozen BERT embeddings capture}: BERT layers encode a hierarchy of linguistic information, with surface features at the bottom, syntactic features in the middle, and semantic features at the top~\citep{jawahar-etal-2019-bert}. Further,~\citet{rogers-etal-2020-primer} show that BERT struggles with pragmatic inference, abstract attributes, and role-based event knowledge. For scene salience prediction, BERT-embeddings complement a subset of our features (character count, abstractness) while subsuming the impact of other features (word order, discourse cues). 

\paragraph{(P3) The LGSM architecture and downstream gains.}
Adding the linguistic stream to a frozen BERT encoder lifts salient-class recall from 51.56\% to 68.34\%, with the gain concentrated on structurally critical scenes whose surface form does not match the conventional semantic profile of salience (Table~\ref{tab:error_comparison}). The precision/recall trade-off is the right one for downstream reasoning: the +3.67 ROUGE-1 gain with stable BERTScore in summarisation (Appendix~\ref{app:summarization_results}) shows the recovered scenes carry signal rather than noise. LGSM also avoids the mid-narrative recall collapse seen in the semantic baseline (Section \ref{sec:arc_gating}), partially mitigating the Lost-in-the-Middle phenomenon \citep{Liu2024}. This supports the introduction's news-vs-long-form contrast: news systems exploit lead-bias and centrality priors \citep{See2017,Mihalcea2004} that long-form fiction lacks, so explicit structural cues are required to recover scenes a semantic-only model misses.

\paragraph{(N1) Scene-aggregate UID carries no independent signal.}
At the scene-aggregate level, the conventional UID summaries (variance, $\Delta$-UID, upper-tail percentile per scene) carry no signal once positional and length controls are present: the $\ell_1$ selection step in the corpus-level regression (\S\ref{sec:corpus_study}) zeros every BERT aggregate, including the variance-based operationalisation that has dominated empirical UID work. This is a granularity result rather than a null, complemented by (P1) above. A parallel large-scale negative result is reported by \citet{juzek2024signal}: neither $\Delta$-UID nor $\mu$-UID outperforms random baselines on syntactic choice across 936 English sentence pairs. Other works also show that symmetric variance and $\Delta$-UID operationalisations \citep{meister-etal-2021-revisiting,uid-word-order-2023} cannot capture this asymmetry. Thus UID, as conventionally aggregated, is an incomplete description of how information density is organised in production. 

\paragraph{(N2) RST and semantic-grounding features are precision filters.}
Removing either from LGSM improves recall but degrades precision (Section \ref{sec:rst_utility}): they suppress false positives on the non-salient class without adding the structurally critical scenes LGSM recovers. \citet{tsipidi2024surprise} models hierarchical information distribution in long-form text with RST, but its utility for salience detection is strictly local here. RST reads a scene's internal rhetorical shape rather than its external plot connectivity, leaving it open to ``structural mimicry'', overvaluing transitional scenes with climax-like formatting; semantic features similarly overvalue concrete filler. Both capture \emph{how} a scene is written, not \emph{why it matters} (Appendix~\ref{app:qualitative_analysis}).

\section{Conclusion}\label{sec:conclusions}
For the task of salience detection in movie screenplays, our experiments shows that UID helps only at the token level, where salient scenes use fewer of the rarest, hardest-to-predict words. Salient scenes are characterized by a moderate cast of central characters, strong third-person reference, simpler sentences, and more abstract language about motivation and conflict. So in long-form narrative, salience comes from how information is paced within a scene and where the scene sits in the story, not from any single surprising word. Our Linguistically-Grounded Salience Model puts this idea to work by combining these explicit features with a frozen BERT encoder through an adaptive gate. It raises recall on salient scenes by 16.78 points and improves downstream summaries by +3.67 ROUGE-1 with a stable BERTScore. The wider lesson is simple: pair neural representations with explicit cognitive and discourse features rather than expecting opaque embeddings to capture both.

\pagebreak


\section*{Limitations}

While LGSM successfully grounds salience in interpretable linguistic features, we acknowledge two primary limitations in our current modeling paradigm. First, from a computational linguistics perspective, our model infers structure implicitly through distributional features but lacks an explicit representation of \textbf{causal event chains}. We do not currently model the logical dependencies between scenes (e.g., $Scene_A \rightarrow Scene_B$), which limits the model's reasoning capabilities in highly non-linear or puzzle-like narratives where deep causal logic outweighs surface salience. Second, from a cognitive modeling standpoint, our framework relies on surprisal to approximate \textbf{bottom-up} attention (reaction to stimuli). We do not yet model \textbf{top-down accessibility} i.e. the viewer's goal-directed attention or varying interest priors. Relatedly, surprisal is a \textbf{retrospective} measure and the natural quantity for our first research question, since Uniform Information Density is a hypothesis about how surprisal is distributed \citep{levy2007speakers, jaeger2010redundancy}. A \textbf{prospective} alternative is uncertainty reduction, defined over the entropy of upcoming continuations \citep{hale2006uncertainty} and applied to suspense and turning points by \citet{wilmot2020suspense}. Whether prospective uncertainty tracks salience where retrospective surprisal does not is left to future work.

\section*{Ethical Considerations}
\paragraph{Data Privacy and Copyright.}
This work utilizes the MENSA benchmark \citep{saxena2024select}, which comprises movie screenplays and Wikipedia summaries. We acknowledge that screenplays are creative intellectual property protected by copyright. Our use of this data is strictly limited to non-commercial academic research under the principles of fair use . We do not distribute raw screenplay texts; all analysis is performed on feature representations or existing public benchmarks.

\paragraph{Bias and Stereotypes.}
We recognize that commercial cinema historically contains social biases, including gender stereotypes and under-representation of minority groups. Since LGSM learns salience from these narratives, there is a risk that the model may encode these biases. For example, consistently predicting scenes depicting physical conflict (often male-dominated) as more ``salient'' than dialogue-driven character development. Users should interpret the model's output as a reflection of narrative conventions rather than an objective measure of importance.
\\ \\
Finally, we acknowledge the use of AI assistance for literature organization and technical descriptions of feature details, with all content verified by human authors and the usage adheres strictly to the ACL guidelines.

\section*{Acknowledgments}
We are grateful for insightful comments on this work by ARR May 2026 reviewers and Prof Frank Keller's research group at the University of Edinburgh. The final author wishes to acknowledge financial support from the following sources: 1. Seed grant funding from IIIT Hyderabad (No: IIIT/R\&D Office/Seed-Grant/2024-25/007) 2. Extramural funding from the Cognitive Science Research Initiative (CSRI) of the Department of Science and Technology (DST) (project no: DST/CSRI/2024/1139).

\bibliography{bibfile,salience,uid}

\appendix

\section{Corpus-Study Regression Setup}\label{app:corpus_setup}

Table~\ref{tab:cs_setup} reports the full setup for the corpus-level post-LASSO regression analysis described in Section~\ref{sec:corpus_study}.

\begin{table*}[h]
\centering
\small
\begin{tabular}{ll}
\toprule
\textbf{Training} & MENSA val (gold labels) \\
 & Subsample balanced within movies \\
 & (5{,}136 scenes) \\
Movies & 50 (50/50 salient/non-salient pairs) \\
Features (input) & 888 across 32 groups \\
Features (post-pruning) & 636 (drop $|r| > 0.95$) \\
Features (post-$\ell_1$) & 200 (C=0.1) \\
Length control & $\log(\text{num\_words})$, forced \\
Inference & post-LASSO Wald \\
\midrule
\textbf{Test set} & MENSA test (8{,}156 scenes) \\
Test AUC, AP, F1 & 0.69, 0.52, 0.55 \\
McFadden pseudo-$R^2$ & 0.23 \\
\bottomrule
\end{tabular}
\caption{Corpus-study regression setup. Cluster-balanced training on gold validation; held-out evaluation on test. Post-LASSO inference \citep{belloni2013least} admits Wald statistics on the selected features.}
\label{tab:cs_setup}
\end{table*}

Table~\ref{tab:top100} reports the top 100 coefficients from the same regression, ranked by $|z|$. We refer to this table at several points in Section~\ref{sec:corpus_study} when discussing which features surface near the top of the ranking and how their effect sizes compare to the dominant positional features.

\begin{table*}[t]
\centering
\scriptsize
\setlength{\tabcolsep}{2pt}
\renewcommand{\arraystretch}{1.05}
\begin{tabular}{rlrrrc @{\hspace{1em}} rlrrrc}
\toprule
\# & Feature & Coef.\ & $z$ & $p$ & sig & \# & Feature & Coef.\ & $z$ & $p$ & sig \\
\midrule
1 & \texttt{scene\_index\_norm} & $+0.364$ & $+7.88$ & $3.3{\times}10^{-15}$ & *** & 51 & \texttt{gc\_dep\_big\_AUX\_after\_DET} & $-0.090$ & $-1.96$ & 0.050 & * \\
2 & \texttt{pos\_edge\_proximity} & $-0.182$ & $-5.21$ & $1.9{\times}10^{-7}$ & *** & 52 & \texttt{gc\_dep\_big\_AUX\_before\_X} & $-0.080$ & $-1.96$ & 0.051 &  \\
3 & \texttt{centrality\_score} & $+0.235$ & $+5.17$ & $2.4{\times}10^{-7}$ & *** & 53 & \texttt{genre\_drama} & $-0.102$ & $-1.95$ & 0.052 &  \\
4 & \texttt{genre\_history} & $-0.164$ & $-3.59$ & 0.0003 & *** & 54 & \texttt{gpt2\_char\_std\_surprisal} & $-0.082$ & $-1.93$ & 0.054 &  \\
5 & \texttt{exclaim\_rate} & $+0.133$ & $+3.52$ & 0.0004 & *** & 55 & \texttt{rst\_depth} & $+0.215$ & $+1.90$ & 0.057 &  \\
6 & \texttt{gc\_conc\_adj\_mean} & $-0.177$ & $-3.34$ & 0.0008 & *** & 56 & \texttt{gc\_dep\_big\_NOUN\_before\_PRON} & $-0.077$ & $-1.90$ & 0.058 &  \\
7 & \texttt{gc\_dep\_big\_ADV\_after\_PRON} & $-0.122$ & $-3.33$ & 0.0009 & *** & 57 & \texttt{gc\_dep\_big\_PROPN\_before\_ADV} & $-0.065$ & $-1.90$ & 0.058 &  \\
8 & \texttt{gc\_present\_tense\_count} & $+0.374$ & $+3.17$ & 0.0015 & ** & 58 & \texttt{gc\_dep\_big\_INTJ\_before\_NOUN} & $+0.065$ & $+1.89$ & 0.058 &  \\
9 & \texttt{gc\_dep\_rel\_list} & $-0.139$ & $-3.14$ & 0.0017 & ** & 59 & \texttt{gc\_pronoun\_third\_plural\_ratio} & $-0.069$ & $-1.89$ & 0.059 &  \\
10 & \texttt{gc\_dep\_big\_ADP\_after\_VERB} & $-0.111$ & $-2.89$ & 0.0039 & ** & 60 & \texttt{gc\_syn\_comp\_isc\_mean} & $+0.110$ & $+1.89$ & 0.059 &  \\
11 & \texttt{gc\_dep\_rel\_discourse} & $-0.147$ & $-2.84$ & 0.0045 & ** & 61 & \texttt{gc\_dep\_rel\_iobj} & $+0.067$ & $+1.88$ & 0.060 &  \\
12 & \texttt{gc\_dep\_big\_PRON\_before\_AUX} & $+0.113$ & $+2.82$ & 0.0048 & ** & 62 & \texttt{gc\_conc\_verb\_std} & $-0.097$ & $-1.88$ & 0.061 &  \\
13 & \texttt{genre\_sport} & $-0.109$ & $-2.79$ & 0.0053 & ** & 63 & \texttt{gc\_dep\_big\_DET\_before\_NUM} & $-0.075$ & $-1.85$ & 0.064 &  \\
14 & \texttt{gc\_dep\_big\_PUNCT\_before\_X} & $-0.182$ & $-2.79$ & 0.0053 & ** & 64 & \texttt{gc\_dep\_big\_CCONJ\_before\_ADV} & $-0.066$ & $-1.83$ & 0.067 &  \\
15 & \texttt{gc\_dep\_rel\_nsubj} & $-0.174$ & $-2.67$ & 0.0076 & ** & 65 & \texttt{gc\_dep\_big\_NOUN\_before\_ADV} & $+0.059$ & $+1.80$ & 0.072 &  \\
16 & \texttt{gc\_dep\_big\_ADV\_after\_ADV} & $+0.096$ & $+2.66$ & 0.0079 & ** & 66 & \texttt{gc\_dep\_big\_NOUN\_before\_PROPN} & $+0.065$ & $+1.80$ & 0.072 &  \\
17 & \texttt{gc\_dep\_big\_ADP\_after\_ADP} & $-0.098$ & $-2.64$ & 0.0082 & ** & 67 & \texttt{surprisal\_slope} & $-0.098$ & $-1.77$ & 0.077 &  \\
18 & \texttt{gc\_question\_count} & $-0.231$ & $-2.62$ & 0.0088 & ** & 68 & \texttt{gc\_conc\_coverage} & $+0.109$ & $+1.76$ & 0.079 &  \\
19 & \texttt{gc\_dep\_big\_PUNCT\_after\_PRON} & $+0.097$ & $+2.61$ & 0.0091 & ** & 69 & \texttt{gc\_dep\_big\_PART\_after\_PRON} & $+0.065$ & $+1.76$ & 0.079 &  \\
20 & \texttt{gc\_dep\_big\_PUNCT\_after\_NUM} & $-0.115$ & $-2.58$ & 0.0098 & ** & 70 & \texttt{dialogue\_ratio} & $-0.085$ & $-1.74$ & 0.081 &  \\
21 & \texttt{gc\_dep\_big\_PRON\_after\_PRON} & $+0.085$ & $+2.43$ & 0.015 & * & 71 & \texttt{gc\_dep\_big\_ADV\_before\_NUM} & $-0.070$ & $-1.74$ & 0.081 &  \\
22 & \texttt{gc\_dep\_rel\_compound} & $-0.136$ & $-2.40$ & 0.017 & * & 72 & \texttt{gc\_dep\_big\_NOUN\_before\_VERB} & $+0.072$ & $+1.74$ & 0.081 &  \\
23 & \texttt{gc\_dep\_big\_INTJ\_after\_INTJ} & $+0.081$ & $+2.36$ & 0.018 & * & 73 & \texttt{smog\_index} & $+0.131$ & $+1.74$ & 0.081 &  \\
24 & \texttt{callback\_to\_ending} & $+0.110$ & $+2.36$ & 0.018 & * & 74 & \texttt{gc\_dep\_big\_ADV\_before\_PROPN} & $-0.065$ & $-1.74$ & 0.081 &  \\
25 & \texttt{gc\_dep\_big\_ADP\_before\_ADP} & $+0.091$ & $+2.35$ & 0.019 & * & 75 & \texttt{gc\_dep\_big\_ADP\_before\_SCONJ} & $-0.381$ & $-1.74$ & 0.082 &  \\
26 & \texttt{log\_num\_words} & $+0.479$ & $+2.33$ & 0.020 & * & 76 & \texttt{gc\_dep\_big\_SCONJ\_before\_ADP} & $-0.119$ & $-1.74$ & 0.082 &  \\
27 & \texttt{gc\_dep\_big\_PROPN\_before\_DET} & $-0.176$ & $-2.32$ & 0.020 & * & 77 & \texttt{gc\_dep\_big\_ADP\_before\_PART} & $-0.057$ & $-1.74$ & 0.083 &  \\
28 & \texttt{gc\_dep\_big\_PART\_after\_NUM} & $+0.076$ & $+2.32$ & 0.021 & * & 78 & \texttt{gc\_dep\_big\_INTJ\_after\_PROPN} & $+0.058$ & $+1.73$ & 0.084 &  \\
29 & \texttt{gc\_dep\_big\_NUM\_before\_ADJ} & $+0.077$ & $+2.31$ & 0.021 & * & 79 & \texttt{flesch\_reading\_ease} & $+0.143$ & $+1.72$ & 0.086 &  \\
30 & \texttt{gc\_dep\_big\_CCONJ\_after\_ADJ} & $-0.085$ & $-2.29$ & 0.022 & * & 80 & \texttt{gc\_dep\_big\_ADV\_after\_INTJ} & $-0.234$ & $-1.72$ & 0.086 &  \\
31 & \texttt{overlap\_prev} & $+0.087$ & $+2.25$ & 0.025 & * & 81 & \texttt{gc\_dep\_big\_INTJ\_after\_AUX} & $+0.074$ & $+1.71$ & 0.088 &  \\
32 & \texttt{gc\_dep\_big\_SYM\_after\_NOUN} & $+0.123$ & $+2.24$ & 0.025 & * & 82 & \texttt{gc\_dep\_big\_ADP\_after\_ADV} & $+0.055$ & $+1.70$ & 0.090 &  \\
33 & \texttt{gc\_conc\_verb\_mean} & $+0.119$ & $+2.21$ & 0.027 & * & 83 & \texttt{gc\_dep\_big\_PRON\_after\_ADV} & $+0.058$ & $+1.69$ & 0.092 &  \\
34 & \texttt{gc\_dep\_rel\_nmod} & $-0.101$ & $-2.21$ & 0.027 & * & 84 & \texttt{gc\_dep\_rel\_cop} & $+0.083$ & $+1.67$ & 0.095 &  \\
35 & \texttt{gc\_dep\_big\_DET\_before\_PROPN} & $+0.089$ & $+2.20$ & 0.028 & * & 85 & \texttt{gc\_pronoun\_third\_singular\_ratio} & $-0.076$ & $-1.67$ & 0.096 &  \\
36 & \texttt{gc\_dep\_rel\_det} & $-0.123$ & $-2.19$ & 0.029 & * & 86 & \texttt{gc\_dep\_big\_PROPN\_after\_PROPN} & $+0.080$ & $+1.66$ & 0.098 &  \\
37 & \texttt{rst\_rel\_explanation} & $-0.083$ & $-2.18$ & 0.030 & * & 87 & \texttt{gc\_question\_sentence\_ratio} & $-0.089$ & $-1.64$ & 0.101 &  \\
38 & \texttt{gc\_dep\_big\_PROPN\_before\_AUX} & $+0.083$ & $+2.17$ & 0.030 & * & 88 & \texttt{ngram2\_char\_min\_surprisal} & $-0.093$ & $-1.64$ & 0.102 &  \\
39 & \texttt{gc\_dep\_big\_PUNCT\_after\_X} & $+0.170$ & $+2.15$ & 0.032 & * & 89 & \texttt{gc\_dep\_big\_INTJ\_after\_ADJ} & $+0.050$ & $+1.64$ & 0.102 &  \\
40 & \texttt{gc\_dep\_big\_CCONJ\_before\_INTJ} & $-0.121$ & $-2.15$ & 0.032 & * & 90 & \texttt{gc\_dep\_big\_ADP\_before\_INTJ} & $-0.085$ & $-1.63$ & 0.104 &  \\
41 & \texttt{ngram2\_char\_median\_surprisal} & $+0.111$ & $+2.15$ & 0.032 & * & 91 & \texttt{gc\_dep\_big\_NUM\_before\_NOUN} & $+0.056$ & $+1.60$ & 0.110 &  \\
42 & \texttt{overlap\_next} & $+0.081$ & $+2.13$ & 0.033 & * & 92 & \texttt{gc\_dep\_big\_ADP\_before\_PRON} & $+0.061$ & $+1.60$ & 0.110 &  \\
43 & \texttt{gc\_dep\_big\_ADJ\_after\_ADP} & $+0.068$ & $+2.11$ & 0.035 & * & 93 & \texttt{gc\_dep\_big\_ADJ\_before\_PROPN} & $-0.072$ & $-1.60$ & 0.111 &  \\
44 & \texttt{gc\_dep\_big\_CCONJ\_before\_PRON} & $-0.071$ & $-2.10$ & 0.036 & * & 94 & \texttt{gc\_dep\_big\_NOUN\_after\_NUM} & $-0.060$ & $-1.59$ & 0.112 &  \\
45 & \texttt{rst\_rel\_attribution} & $-0.094$ & $-2.08$ & 0.038 & * & 95 & \texttt{gc\_dep\_big\_PRON\_before\_PART} & $-0.075$ & $-1.58$ & 0.114 &  \\
46 & \texttt{genre\_biography} & $+0.097$ & $+2.07$ & 0.039 & * & 96 & \texttt{gc\_dep\_big\_PROPN\_before\_PROPN} & $+0.078$ & $+1.58$ & 0.115 &  \\
47 & \texttt{gc\_dep\_big\_NUM\_before\_VERB} & $+0.074$ & $+2.02$ & 0.043 & * & 97 & \texttt{gc\_dep\_big\_PART\_before\_ADJ} & $+0.053$ & $+1.57$ & 0.116 &  \\
48 & \texttt{gc\_dep\_big\_PUNCT\_after\_AUX} & $-0.123$ & $-2.00$ & 0.045 & * & 98 & \texttt{gc\_dep\_big\_PUNCT\_before\_NUM} & $+0.065$ & $+1.56$ & 0.118 &  \\
49 & \texttt{gc\_dep\_big\_VERB\_after\_NOUN} & $-0.094$ & $-1.98$ & 0.048 & * & 99 & \texttt{gc\_cd\_mattr} & $+0.111$ & $+1.56$ & 0.118 &  \\
50 & \texttt{gc\_dep\_big\_PUNCT\_before\_ADJ} & $-0.078$ & $-1.97$ & 0.049 & * & 100 & \texttt{gc\_dep\_big\_DET\_before\_ADP} & $-0.074$ & $-1.53$ & 0.125 &  \\
\bottomrule
\end{tabular}
\caption{Top 100 coefficients from the post-LASSO regression on the MENSA gold validation split, ranked by $|z|$. Coefficients are standardised (features were z-scored before fitting), so magnitudes are directly comparable. Stars indicate per-coefficient significance levels: *** $p<0.001$, ** $p<0.01$, * $p<0.05$. The three top features (rows 1--3) are positional or structural and are well separated from the rest of the distribution by $|z|$. Length is force-included as $\log(\text{num\_words})$; all coefficients are net of length and within-movie position.}
\label{tab:top100}
\end{table*}

\section{Implementation Details}\label{app:implementation}

We implemented the LGSM framework using PyTorch and the HuggingFace Transformers library.

\subsection{Model Architecture}

\paragraph{Semantic Encoder:} We utilized the frozen \texttt{bert-large-uncased} model ($L = 24$ layers, $H = 1024$ hidden size, $A = 16$ attention heads) to extract scene embeddings. The encoder parameters were frozen during training to reduce computational cost and prevent overfitting.

\paragraph{Linguistic Projector:} The linguistic feature vector $\mathbf{x}_{\text{ling}} \in \mathbb{R}^k$ is projected independently via a 1-layer MLP with LayerNorm, GELU activation, and a dropout rate of 0.1, giving $\mathbf{h}_{\text{ling}} \in \mathbb{R}^{d_{\text{ling}}}$. Within the Adaptive Gated Fusion step, a second linear map $W_l$ projects $\mathbf{h}_{\text{ling}}$ into the shared $d_{\text{model}} = 1024$ space before it is combined with the semantic stream via the scalar gate $\alpha_t$; the two streams are never concatenated.

\paragraph{Contextual Encoder:} The Transformer Encoder consists of $L = 2$ layers (reduced from standard BERT to prevent overfitting on small batch sizes) with 16 attention heads and a feed-forward dimension of 4096 (4$\times$ the hidden dimension). We employ GELU activation and dropout of 0.1 throughout the encoder.

\paragraph{Positional Encoding:} We add sinusoidal positional encodings to the fused features before the sequence model to capture the temporal ordering of scenes within each movie.

\paragraph{Classifier:} The final classification head uses a linear layer that takes as input the concatenation of the transformer output (contextual features) and the original fused features (local features) via a skip connection, resulting in an input dimension of 2048.

\subsection{Optimization}

We trained the model using the AdamW optimizer with a learning rate of $2 \times 10^{-5}$ for all parameters, weight decay of 0.01, and gradient clipping with a maximum norm of 1.0. We employed a linear warmup for the first 10\% of training steps (if warmup is enabled), followed by a ReduceLROnPlateau scheduler that reduces the learning rate by a factor of 0.5 when validation performance plateaus, with a patience of 3 epochs and a minimum learning rate of $10^{-6}$.

\subsection{Feature Processing}

Linguistic features were normalized and clipped to the range $[-10, 10]$ to prevent numerical instability. Scene embeddings were L2-normalized before fusion.

\subsection{Focal Loss Configuration}
For the focal loss objective, we performed a grid search on the validation set for the focusing parameter $\gamma \in \{1.0, 2.0, 3.0, 5.0\}$ and the balance parameter $\alpha \in \{0.25, 0.5, 0.75\}$. The optimal performance was achieved with $\gamma=3.0$ and $\alpha=0.25$. Here $\gamma$ down-weights easy, well-classified examples of both classes, concentrating the loss on hard cases; $\alpha=0.25$ additionally re-weights the loss toward the minority salient class, consistent with our recall-oriented objective.

\section{Baseline Implementation Details}
\label{app:baselines}

\paragraph{Weighted Ensemble.}
The machine learning ensemble described in Section~\ref{sec:results} consists of the following 13 classifiers implementation using \texttt{scikit-learn}:
\begin{itemize}
    \item \textbf{Linear Models:} Logistic Regression ($C=1.0$), SGD Classifier, Passive Aggressive Classifier.
    \item \textbf{Support Vector Machines:} Linear SVC, RBF SVC, Polynomial SVC, Sigmoid SVC.
    \item \textbf{Tree-Based:} Random Forest ($n=100$), Extra Trees, Decision Tree.
    \item \textbf{Neighbors/others:} K-Neighbors ($k=5$), Gaussian NB, AdaBoost.
\end{itemize}
The final prediction is a soft-voting ensemble where the contribution of each classifier is weighted by its standalone F1 score on the validation set.

\paragraph{Zero-shot generative baselines.}
The Qwen and Gemini 2.5 Flash rows in Table~\ref{tab:classification_results} are
zero-shot. Each scene is rated for salience against a single fixed definition,
with no in-context examples. Each scene is scored on a $0$ to $1$ salience scale
and the scores are converted to binary salient / non-salient labels. The full
prompt appears below.

\begin{quote}
\small\ttfamily\raggedright
\textbf{[System]}\\
You are an expert film analyst rating scenes in a movie screenplay for narrative
salience. Narrative salience is the likelihood that a scene would be mentioned
if someone gave a concise plot summary of the film: the probability that a
reader retelling the story would include what happens in that scene. Rate the
scene with a float strictly between 0.0 and 1.0, where higher means more likely
to appear in a plot summary. Salient scenes carry the main storyline forward
(turning points, central conflicts introduced or resolved, consequential plot
developments). Non-salient scenes are background, mood, transitions, or minor
exposition a summary would omit, even if vivid, intense, or action-heavy. Judge
each scene by its function in the whole story, not its surface drama. Respond
only with the score, a single float in (0.0, 1.0).

\medskip
\textbf{[User]}\\
MOVIE: \{name\}\\
SCENE: \{scene\_text\}
\end{quote}

\section{Feature Specifications}
\label{app:feature_specs}

This appendix provides the operational specifications for the feature groups described in Section \S\ref{sect:features}. Unless otherwise noted, we extract distributional statistics (mean, variance, maximum, minimum) aggregated at the scene level.

\subsection{Scene Length}
\label{app:length}

We extract three length measures at the scene level. \textbf{Raw token count} is computed after whitespace and punctuation normalisation, excluding scene-heading boilerplate (slug lines, transition cues such as \texttt{CUT TO:}) so that the measure reflects the body of the scene rather than formatting overhead. \textbf{Log word count}, $\log(\text{num\_words})$, is the same quantity on a log scale; this is the form used as a forced covariate in the corpus-level regression (\S\ref{sec:corpus_study}, Appendix~\ref{app:corpus_setup}) to absorb length-driven variance before testing other predictors. \textbf{Sentence count} is obtained from a sentence segmenter applied to action and dialogue blocks jointly, providing a granularity-controlled complement to the token-level measures: two scenes with the same token count but differing sentence counts differ in pacing and turn structure.

\subsection{Surprisal and Cognitive Processing}
\label{app:surprisal}

We operationalize cognitive processing dynamics through six complementary streams, organized by model architecture. First, we model \textbf{causal surprisal} using the pre-trained GPT-2 architecture. We compute character-level surprisal by tokenizing text into characters to capture orthographic friction distinct from semantics, and we extract dynamic trajectory features by calculating the linear slope of token-level surprisal across the scene. Second, we measure \textbf{semantic fit and cognitive load} using BERT-large. We compute Pseudo-Log-Likelihood (PLL) scores by iteratively masking tokens to measure contextual coherence, and we estimate sustained load using the PsychFormers toolkit\footnote{\url{https://github.com/jmichaelov/PsychFormers}} to process scenes in maximal contiguous chunks (512 tokens), summing sequential PLL scores. Third, we establish \textbf{statistical baselines} using KenLM n-gram models (orders 1--5) trained on a screenplay corpus. We compute separate streams for word n-grams (isolating lexical rarity) and character n-grams (quantifying visual loudness and formatting anomalies).

\subsection{Morphosyntactic Structure}
This feature group captures grammatical scaffolding across three dimensions. First, we model \textbf{syntactic complexity} via dependency-based representations, extracting the argument–adjunct ratio, tree depth, dependency distance, and complexity indices. We also extract dependency bigrams to encode recurrent local syntactic patterns. Second, we capture \textbf{grammatical style} through part-of-speech distributions, normalizing counts of nouns, verbs, and adjectives. Third, we operationalize \textbf{temporal anchoring} by counting past-tense verbs, present-tense verbs, and temporal adverbs, using tense dominance ratios as proxies for temporal structure.

\subsection{Lexical Specificity and Stylistic Register}
We measure type-token ratios, character-level diversity indices, and screenplay-specific patterns (action verb density, dialogue markers) \cite{elson2010extracting}. Diversity is computed locally at the scene level to capture informational contribution rather than document-wide richness.

\subsection{Semantic Grounding and Register}
We capture semantic texture using a dual-stream approach. To model \textbf{semantic grounding}, we extract concreteness features from the Brysbaert et al. (2014) norms. We compute scene-level means - optionally stratified by part-of-speech (nouns, verbs, adjectives) and joint distributions with sentiment to distinguish grounded action from abstract dialogue. Simultaneously, we track \textbf{stylistic register} by calculating the density of academic and formal vocabulary, capturing shifts in lexical formality.

\subsection{Discourse-Pragmatic Structure}
\begin{figure}[h]
    \centering
    \includegraphics[width=\linewidth]{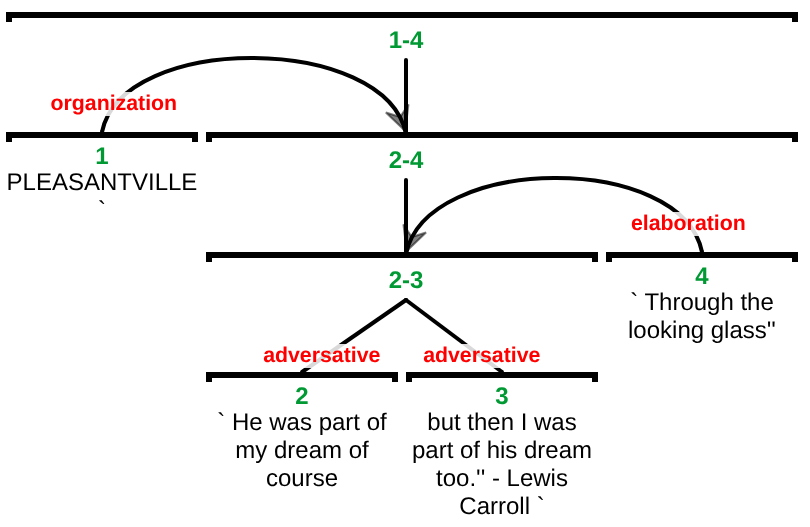}
    \caption{Neural RST parse tree for a sample scene, illustrating the hierarchical structure used to model discourse depth.}
    \label{fig:rst_tree}
\end{figure}

We operationalize discourse organization through three methods. First, we apply \textbf{RST parsing} to extract structural properties from rhetorical trees, utilizing tree depth and maximum path length as proxies for argumentative complexity. Second, we compute \textbf{TextRank centrality} on a scene-level graph weighted by cosine similarity of semantic embeddings, modeling plot connectivity. Third, we track \textbf{surface discourse markers}, counting the density of explicit causal, contrastive, and additive connectives. Figure~\ref{fig:rst_tree} shows an example parse.

\subsection{Dialogic Interaction and Deixis}
We characterize interactional dynamics by tracking the balance between speech and description. 
We track dialogue density, speech-verb usage, and pronoun distributions. Pronoun statistics capture referential focus shifts, distinguishing scenes driven by direct interpersonal engagement from those characterized by detached narration.

\subsection{Emotional and Evaluative Dynamics}
We extract sentiment polarity via distributional lexicons and model affective dynamics through categorical emotion measures and trajectory features (variance and slope) to distinguish volatile affective shifts from consistent emotional tones.

\subsection{Narrative-Structural Dynamics}
We capture global structure through four lenses. First, \textbf{screenplay structure} is modeled via normalized scene position and three-phase segmentation bins. Second, \textbf{plot shifts} are quantified using lexical overlap with neighboring scenes, vocabulary novelty, similarity change magnitude, and dialogue turnover. Third, \textbf{thematic callbacks} are measured via the lexical similarity of the current scene to the movie's opening and ending. Fourth, \textbf{character-centric progression} is tracked using named-entity recognition to compute dynamic metrics: character turnover rates, introductions, and reintroductions after absence.

\subsection{Baseline Salience Priors}
We incorporate baseline predictions by extracting salience scores from the hierarchical memory network model as a single continuous feature. This enables residual learning, allowing the model to refine existing semantic estimates by incorporating explicit structural priors.

\subsection{Contextual and Genre Conditioning}
We condition salience predictions on generic context by encoding film genre indicators as dense feature vectors. We utilize multi-hot encoding for major categories (Action, Drama, Thriller, Comedy, Romance, Horror, Science Fiction). These vectors allow the classifier to dynamically adapt feature weighting, for example, upweighting verb density for action films versus dialogue density for dramas, ensuring the representation of importance remains context-sensitive.

\section{Feature Statistics}
\label{app:feature_stats}

\begin{table*}[t]
\centering
\small
\begin{tabular}{@{}lrrrcc@{}}
\toprule
\textbf{Feature} & \textbf{Salient} & \textbf{Non-Salient} & \textbf{Diff.} & \textbf{Cohen's $d$} & \textbf{Sig.} \\
\midrule
BERT Variance Surprisal & 8.34 & 10.31 & $-1.97$ & $-0.41$ & *** \\
Central Character Count & 5.66 & 2.48 & $+3.18$ & $+0.64$ & *** \\
Highly Abstract Word Count & 27.93 & 10.32 & $+17.61$ & $+0.78$ & *** \\
Unique Person Count & 4.73 & 2.17 & $+2.56$ & $+0.75$ & *** \\
BERT Q90 Surprisal & 5.71 & 6.54 & $-0.83$ & $-0.45$ & *** \\
Flesch--Kincaid Grade & 4.69 & 5.30 & $-0.61$ & $-0.23$ & *** \\
Dialogue Count & 4.81 & 1.73 & $+3.08$ & $+0.69$ & *** \\
Gunning Fog Index & 6.49 & 7.03 & $-0.54$ & $-0.18$ & *** \\
Third-Person Singular Pronoun & 12.83 & 4.96 & $+7.87$ & $+0.74$ & *** \\
Automated Readability Index & 4.89 & 5.61 & $-0.72$ & $-0.24$ & *** \\
\bottomrule
\end{tabular}
\caption{Mean feature values and effect sizes for salient and non-salient scenes on the MENSA \textbf{validation set} (8052 scenes from 50 movies), showing the top 10 features ranked by mean absolute SHAP importance. All differences are statistically significant ($p < 0.001$, denoted by ***).}
\label{tab:feature_averages}
\end{table*}

Table~\ref{tab:feature_averages} presents the mean values of the top ten most influential features (ranked by mean absolute SHAP contribution) for salient and non-salient scenes. Several consistent patterns emerge that directly inform the interpretation in Section~\ref{sec:shap_analysis}.

\paragraph{Cognitive features.}
Salient scenes exhibit substantially lower surprisal variance ($-1.97$) and lower upper-tail surprisal ($-0.83$), indicating controlled rather than extreme unpredictability. This is consistent with the monotonic decreasing relationship confirmed in Appendix~\ref{app:bert_surprisal}: salience is highest at low surprisal and declines steadily as tail surprisal rises, with no significant curvature ($p = 0.137$).

\paragraph{Character structure.}
Salient scenes contain more central characters ($+3.18$) and more unique persons ($+2.56$). However, validation-set regression analysis (Appendix~\ref{app:nonmonotonic_character_structure}) reveals a non-monotonic effect for central character count, where salience increases up to an intermediate range and then declines. This distinguishes focused character convergence from scenes with diffuse participation.

\paragraph{Stylistic register.}
All three readability metrics - Flesch-Kincaid Grade ($-0.61$), Gunning Fog Index ($-0.54$), and Automated Readability Index ($-0.72$) are lower for salient scenes, confirming a strong preference for syntactic simplicity in narratively important moments. This aligns with psycholinguistic evidence that simpler constructions facilitate rapid integration during high-stakes events.

\paragraph{Dialogic focus.}
Salient scenes exhibit higher dialogue counts ($+3.08$) and markedly higher third-person singular pronoun usage ($+7.87$), capturing sustained referential focus on specific individuals performing consequential actions rather than collective narration or diffuse description.

\paragraph{Semantic grounding.}
Counterintuitively, salient scenes contain substantially more highly abstract words ($+17.61$). This pattern reflects the tendency for narratively pivotal moments to involve abstract concepts, such as motivation, intention, or moral conflict, whereas concrete sensory description more often characterizes background or transitional material.

Taken together, the distributional patterns in Table~\ref{tab:feature_averages} complement the SHAP analysis in Figure~\ref{fig:shap_contributions_appendix} by revealing how salient and non-salient scenes differ in their underlying linguistic profiles. The convergence of mean differences, SHAP correlations, and bidirectional SHAP contributions demonstrates that LGSM learns conditional, context-dependent feature usage rather than simple monotonic heuristics, enabling it to recover structurally critical but semantically atypical scenes that purely semantic baselines systematically miss.

\section{SHAP Attribution Procedure}\label{app:shap_procedure}

This section provides methodological details for our SHAP-based interpretability analysis to ensure reproducibility and address attribution concerns.

\subsection{SHAP Variant and Rationale}

We employ \textbf{KernelSHAP} (Lundberg and Lee, 2017) as our attribution method. KernelSHAP is model-agnostic and provides Shapley values satisfying efficiency, symmetry, dummy, and additivity axioms. We selected KernelSHAP over DeepSHAP because: (1) our architecture combines multiple components (BERT, linguistic features, gated fusion), making model-agnostic attribution more appropriate; (2) we require feature-level (not token-level) explanations; and (3) our feature space is computationally manageable.

\subsection{Background Dataset and Sampling}
\textbf{Background Dataset}: We use a fixed sample of scenes drawn from the MENSA training split as the background dataset for KernelSHAP, maintaining computational efficiency across all evaluated movies.
\\
\textbf{Sampling Strategy}: For each scene explanation:
\begin{itemize}
    \item \textit{Background Samples}: randomly sampled scenes per movie, selected from the movie's training scenes.
    
    \item \textit{L1 Regularization}: We use L1 regularization with top-50 features to improve convergence and focus on the most important features.
\end{itemize}

All SHAP attributions are computed at the scene level, reflecting our binary
scene-level classification task. Each scene receives independent SHAP values
for all features. For movie-level statistics (e.g., gate-prediction correlations),
we aggregate scene-level SHAP values using mean absolute SHAP values per
feature across all scenes in a movie.

\subsection{Handling Correlated Features}

We address feature correlation through the following strategies:

\begin{enumerate}
    \item \textit{Feature Group Aggregation}: We report SHAP values both at the individual feature level and aggregated by feature category (Discourse, BERT, Sentiment, Position, Linguistic, Other). This category-based aggregation reduces the impact of within-category correlations while preserving interpretability.
    
    \item \textit{Feature Value vs. SHAP Correlation}: We compute correlations between feature values and their SHAP contributions to identify non-linear relationships. Features with high absolute correlation (|r| > 0.85) show strong linear relationships with SHAP values, while lower correlations indicate non-linear effects. This analysis helps identify which features have consistent directional effects (e.g., RST depth: r = 0.93 with SHAP) versus those with context-dependent contributions.
\end{enumerate}

\section{Ablation Diagnostics and Qualitative Analysis}
\label{app:qualitative_analysis}

As shown in Table~\ref{tab:discordant_pairs}, both feature groups improve performance primarily by suppressing false positives on non-salient scenes. Table~\ref{tab:qualitative_examples} presents representative success and failure cases illustrating physical grounding, structural mimicry, and abstract suppression effects.

\begin{table*}[h]
\centering
\small
\begin{tabular}{lccccccc}
\toprule
\multirow{2}{*}{\textbf{Ablated Group}} & \multirow{2}{*}{\textbf{$\chi^2$ ($p$-val)}} & \multicolumn{2}{c}{\textbf{Total Discordant}} & \multicolumn{2}{c}{\textbf{Salient}} & \multicolumn{2}{c}{\textbf{Non-Salient}} \\
\cmidrule(lr){3-4} \cmidrule(lr){5-6} \cmidrule(lr){7-8}
 & & \textbf{Helped} & \textbf{Hurt} & \textbf{Help} & \textbf{Hurt} & \textbf{Help} & \textbf{Hurt} \\
\midrule
Discourse (RST) & 56.60 ($<.001$) & 287 & 132 & 25 & 95 & 262 & 37 \\
Semantic (Conc.) & 116.88 ($<.001$) & 473 & 193 & 11 & 182 & 462 & 11 \\
\bottomrule
\end{tabular}%
\caption{Discordant prediction pairs (Full vs. Ablated models)}
\label{tab:discordant_pairs}
\end{table*}

\begin{table*}[h]
\centering
\small
\begin{tabular}{@{}p{1.5cm}p{2.5cm}p{5.5cm}p{5cm}@{}}
\toprule
\textbf{Feature} & \textbf{Mode} & \textbf{Scene Excerpt} & \textbf{Diagnostic Analysis} \\
\midrule
\textbf{RST} & \textbf{Success}\newline (Sequential Pivot) & \textit{Ninja Assassin} (Salient): ``We hear his heartbeat. Bump-thump... As he stands on the edge... In the next moment he is gone - And finally, free.'' & Captures rapid temporal/spatial shifts characteristic of a resolution beat, despite a lack of dialogue. \\
\addlinespace
\textbf{RST} & \textbf{Failure}\newline (Mimicry) & \textit{The Shining} (Non-Salient): ``WENDY: Danny, I can't get out!... WENDY: Run, run and hide. Run, quick!'' & Intense rhetorical structure (rapid commands, intercutting) mimics a climax, deceiving the model into flagging this transitional beat as salient. \\
\midrule
\textbf{Semantic} & \textbf{Success}\newline (Physical Grounding) & \textit{Moonstruck} (Salient): ``Ronny appears and approaches the door. HE KNOCKS, waits, and then RINGS THE BELL. Rose answers...'' & Dense physical verbs (knocks, rings, answers) ground the model, allowing it to correctly identify plot-driving narrative momentum. \\
\addlinespace
\textbf{Semantic} & \textbf{Failure}\newline (Abstract Suppression) & \textit{Bamboozled} (Non-Salient): ``...you will get a guest-starring speaking role... plus an added bonus: a lifetime of Da Bomb 125\% Malt Liquor.'' & Highly abstract constructs (role, bonus, lifetime) suppress the salience score, reflecting a blind spot for scenes driven by structural or societal stakes. \\
\bottomrule
\end{tabular}
\caption{Qualitative examples illustrating the success mechanisms and failure modes of the linguistic features.}
\label{tab:qualitative_examples}
\end{table*}

\section{Narrative-arc Stability and Gating Behaviour}\label{app:arc_gating}

This appendix hosts the figures supporting the narrative-arc and gating analysis in \S\ref{sec:arc_gating}.

\begin{figure}[htbp]
    \centering
    \includegraphics[width=\linewidth]{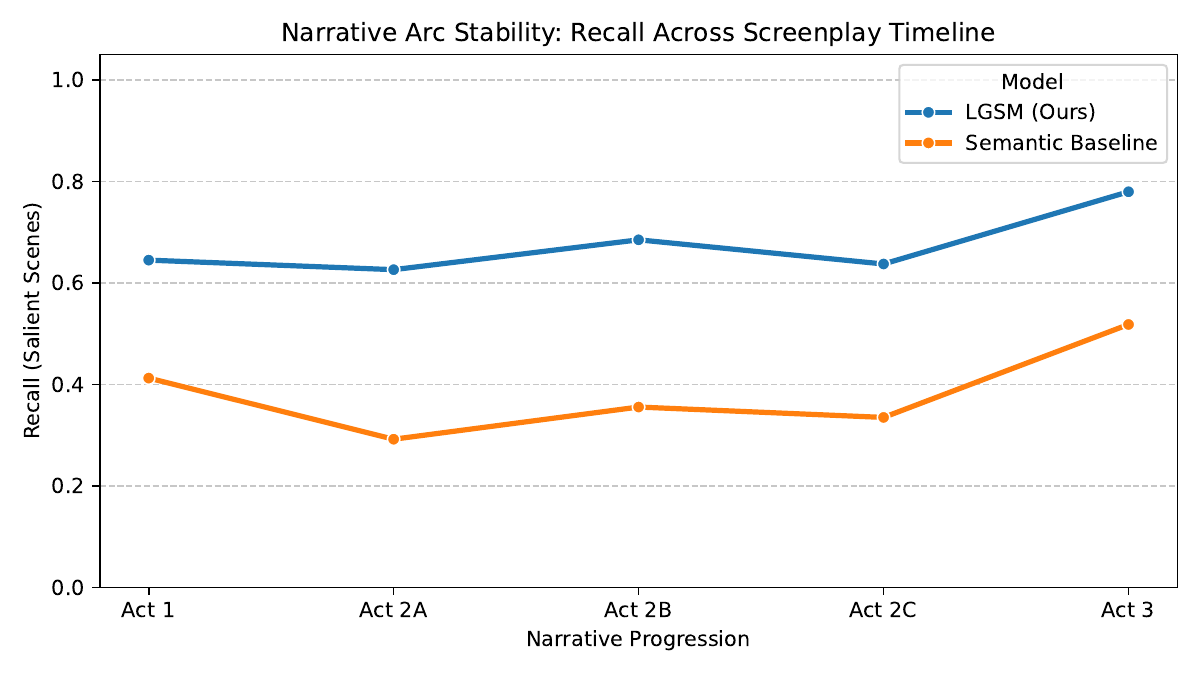}
    \caption{Recall across the narrative timeline (Acts 1--3) for LGSM versus the semantic baseline. The baseline drops from Act 1 to Act 2A and stays low through Act 2C; LGSM remains stable across the entire arc.}
    \label{fig:narrative_arc}
\end{figure}

\section{Gating Behavior Diagnostics}
\label{app:gating_diagnostics}

\subsection{Within-Movie Stability}

\begin{figure*}[h]
    \centering
    \includegraphics[width=0.95\textwidth]{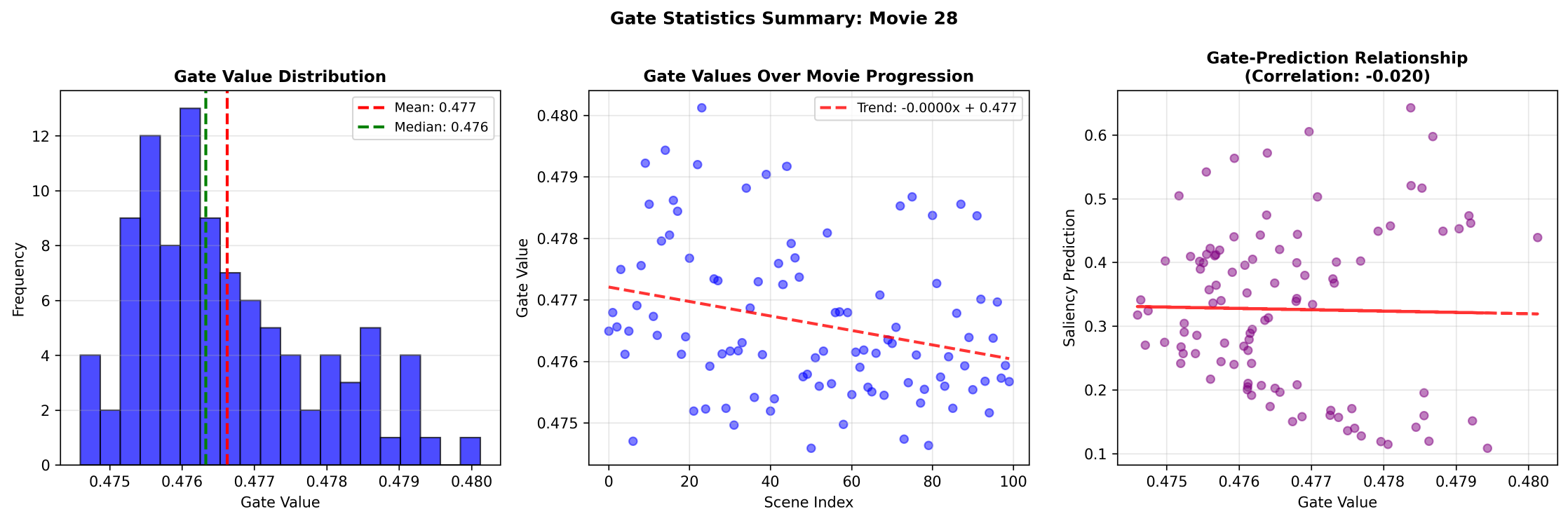}
    \caption{Gate statistics for Movie 28. \textbf{(Left)} Distribution of gate values (Mean=0.477, $\sigma \approx 0$). \textbf{(Center)} Gate values over scene index ($slope \approx 0$). \textbf{(Right)} Gate value vs.\ salience prediction ($r=-0.02$).}
    \label{fig:gate_statistics_movie_28}
\end{figure*}

Figure~\ref{fig:gate_statistics_movie_28} shows the gate values for the movie ``Her'' (Movie 28). The values cluster tightly around $\mu = 0.477$ with very low variance, and remain flat across scene indices ($slope \approx 0$). This means the model gives roughly equal weight to linguistic features and semantic embeddings throughout the entire film, not just in early scenes. The correlation between the gate value $\alpha$ and the predicted salience score is near zero ($r = -0.020$), indicating that the gate controls the balance between the two input streams regardless of whether a scene is predicted as salient or not.

\begin{figure*}[h]
    \centering
    \includegraphics[width=0.95\textwidth]{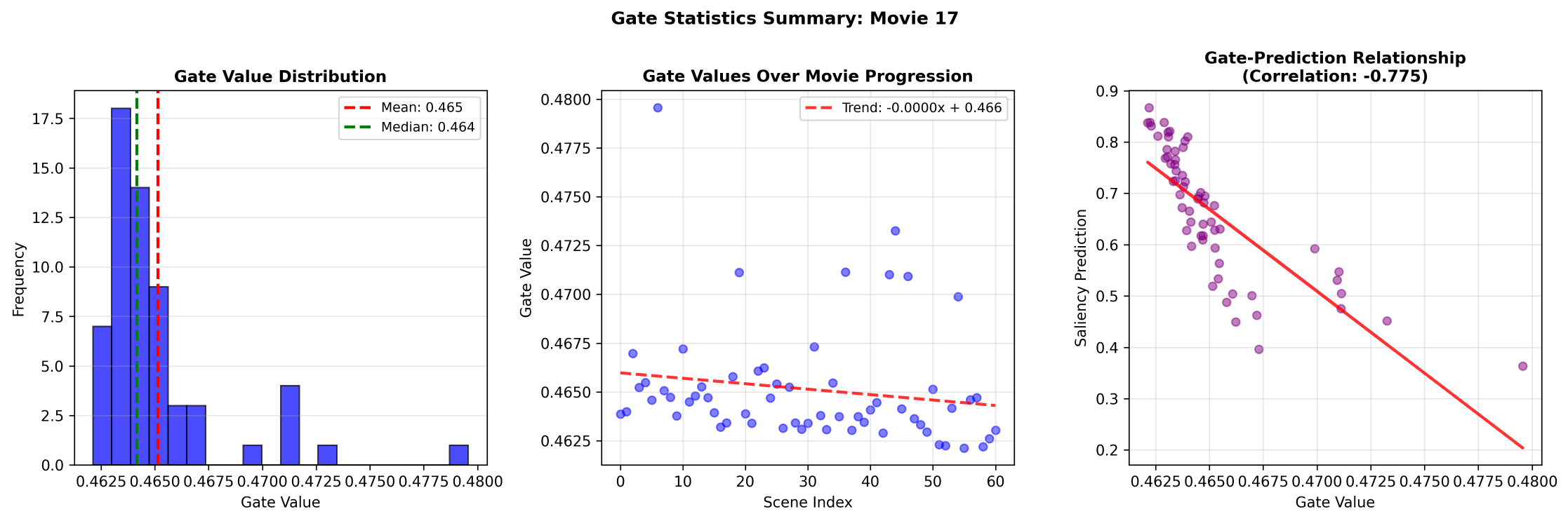}
    \caption{Gate statistics for Movie 17}
    \label{fig:gate_statistics_movie_17}
\end{figure*}

\subsection{Across-Movie Variation}

The gate is stable within a given movie but varies across movies (Figures~\ref{fig:gate_statistics_movie_28} and~\ref{fig:gate_statistics_movie_17}). In both cases the temporal trend is flat (mean $|slope| < 1e{-4}$), but the gate relates to the prediction differently. For most movies the correlation is near zero ($r \approx 0$). However, Movie 17 shows a strong negative correlation ($r = -0.775$), where higher gate values (more weight on linguistic features) correspond to lower salience predictions. This suggests that in some films the model uses linguistic features mainly to identify and reject background scenes, and relies more on semantic embeddings when predicting a scene as salient.

\subsection{Range of Gate Behaviour Across Movies}
The two case studies illustrate the range of gate behaviour observed in LGSM. In Movie 17, the gate value is strongly correlated with the model's salience prediction ($r=-0.775$): scenes where the gate assigns more weight to the linguistic stream are more likely to be predicted non-salient. In Movie 28, the gate is essentially uncorrelated with the prediction ($r=-0.020$) and stays flat across the film ($\mu=0.477$, $\sigma\approx0$), so the two streams are combined at a near-constant ratio regardless of the outcome. These two movies show that the gate--prediction relationship is not fixed across films: it ranges from near-zero to strongly negative.

Separately, at the corpus level, ablating the RST and semantic-grounding feature groups raises recall mainly by removing correctly-rejected non-salient scenes rather than true positives (Table~\ref{tab:discordant_pairs}): of the discordant predictions, the large majority of ``helped'' cases fall on the non-salient side. This shows that, in aggregate, these feature groups act primarily to suppress false positives rather than to detect additional salient scenes. We report this alongside the gate case studies as related evidence on how the linguistic stream is used, without claiming a direct causal link between the two analyses; testing whether they reflect the same underlying mechanism is left to future work.

\section{Non-Monotonic Effects of Character Structure on Scene Salience}
\label{app:nonmonotonic_character_structure}

This appendix reports validation-set analyses examining the robustness of the non-monotonic character-structure effects. The central character count effect (\S\ref{sec:shap_analysis}, Figure~\ref{fig:central_character_nonmonotonic}) is complemented by the unique character count analysis below; both show diminishing returns at higher counts.

\subsection{Unique Character Count}

\begin{figure}[htbp]
    \centering
    \includegraphics[width=0.95\linewidth, keepaspectratio]
    {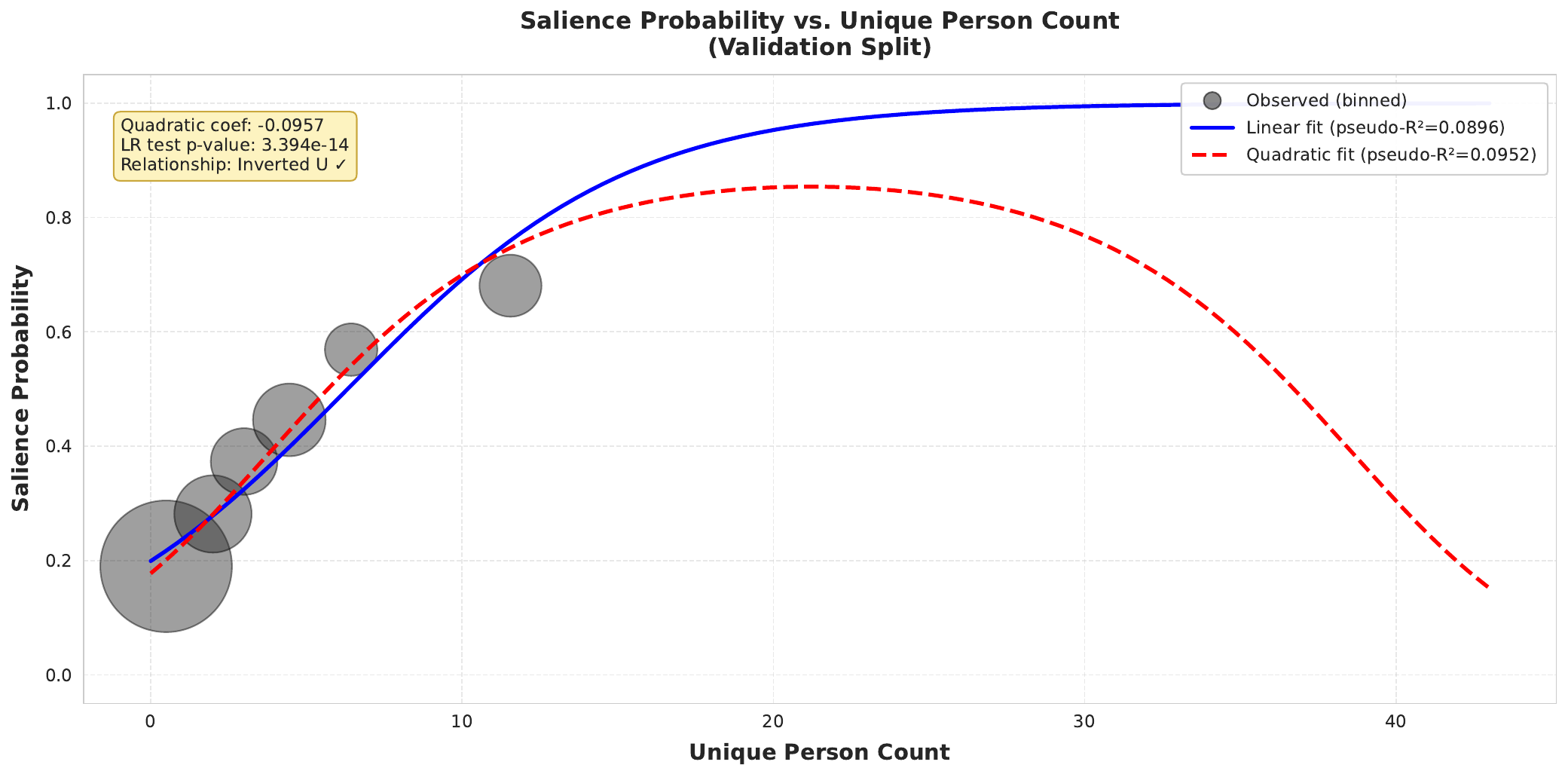}
    \caption{Validation-set salience probability as a function of unique character count. Observed salience rates are shown using binned averages, with linear (dashed) and quadratic (solid) logistic regression fits. The negative quadratic term indicates diminishing returns at high character counts.}
    \label{fig:unique_person_nonmonotonic}
\end{figure}

Figure~\ref{fig:unique_person_nonmonotonic} shows the relationship between \textit{unique character count} and scene salience on the validation split. Salience probability increases from very low values, indicating that scenes involving multiple participants are more likely to be narratively important. However, the quadratic model again yields a statistically significant negative second-order term ($\beta_2 < 0$, $p < 0.001$), indicating diminishing returns at higher character counts. While the absolute gain over a linear baseline remains modest, the effect is stable across splits, confirming that scenes with extremely high character diversity are less likely to be salient despite involving many distinct participants.

\section{Additional Validation Analysis of BERT Surprisal Features}
\label{app:bert_surprisal}

This appendix reports validation-set analyses for the BERT-based surprisal features discussed in Section~\ref{sec:shap_analysis}. While SHAP-based attribution analysis identifies these features as influential in the model's decision process, the following results clarify how their raw values relate to empirical salience probabilities on held-out data. In particular, we distinguish between effects that remain non-linear under validation and those that resolve into stable monotonic trends.

\subsection{Upper-Tail Surprisal}

\begin{figure}[htbp]
    \centering
    \includegraphics[width=0.95\linewidth, keepaspectratio]
   {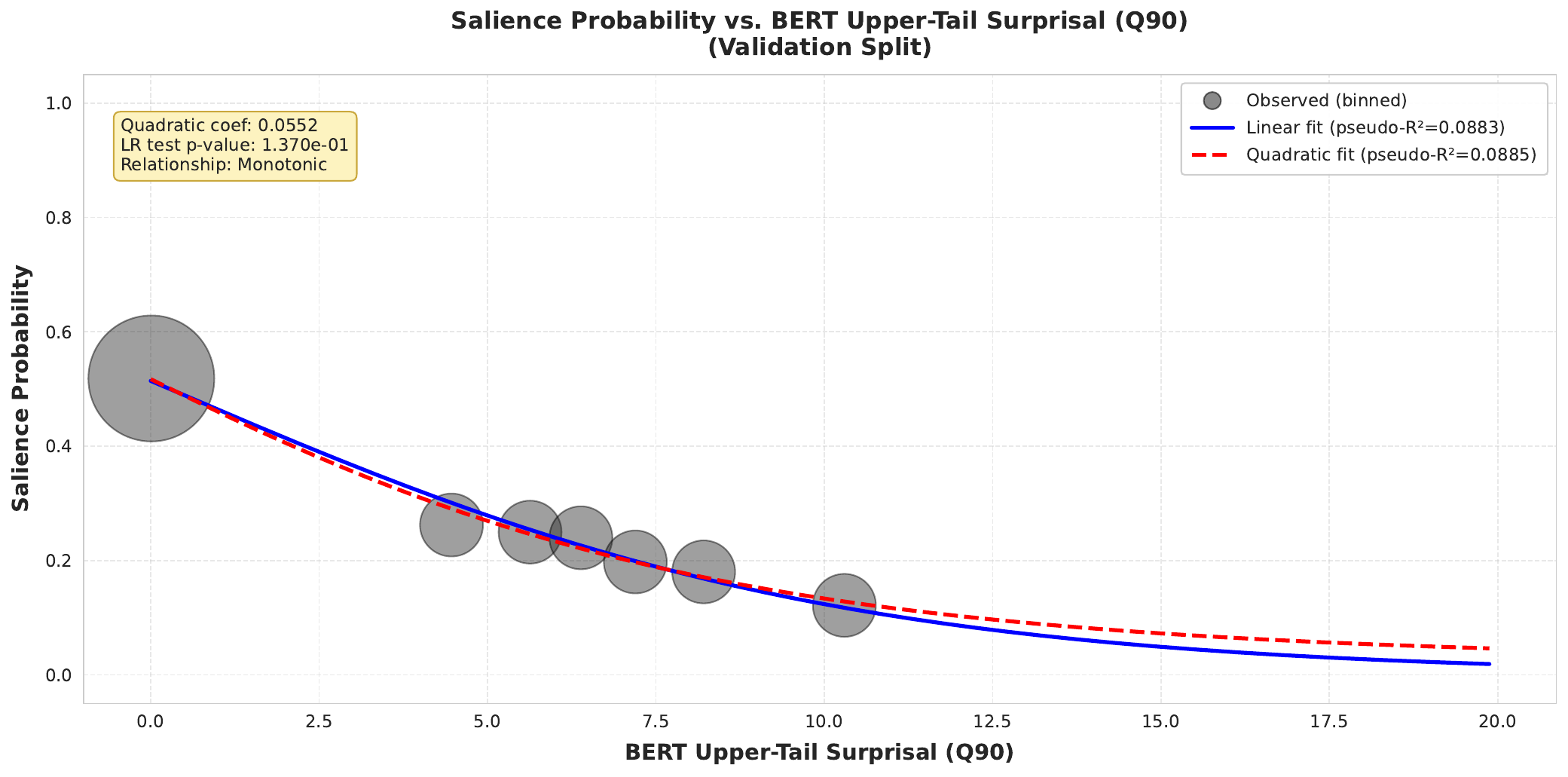}
    \caption{Validation-set salience probability as a function of BERT upper-tail surprisal (90th percentile). Salience decreases monotonically with increasing tail surprisal, and extreme values correspond primarily to non-salient scenes.}
    \label{fig:bert_q90_nonmonotonic}
\end{figure}

Figure~\ref{fig:bert_q90_nonmonotonic} shows validation-set salience probability as a function of \textit{upper-tail surprisal} (90th percentile). Empirical salience rates are computed using binned averages, with linear and quadratic logistic regression fits overlaid. On the validation split, the quadratic term is not statistically significant, and the relationship is well characterized by a monotonic decreasing trend. Salience probability is highest at low surprisal values and declines steadily as tail surprisal increases. Although the quadratic fit exhibits a slight rebound at extreme values, these regions are sparsely populated and dominated by non-salient scenes, indicating that extreme lexical rarity does not reliably signal narrative importance.

\section{Dataset Details and Statistics} \label{app:dataset_details}

\paragraph{Data and Splits.}
MENSA contains 924 movies with 133,711 scenes. The training split (824 movies; 117,503 scenes) uses \textit{silver} labels obtained via automatic script-summary alignment, while validation and test splits (50 movies each) use \textit{gold} human annotations (Table~\ref{tab:mensa_split_stats}).
\begin{table}[t]
\centering
\small
\setlength{\tabcolsep}{6pt}
\begin{tabular}{lrrr}
\toprule
\textbf{Statistic} & \textbf{Train} & \textbf{Val} & \textbf{Test} \\
\midrule
\#Movies & 824 & 50 & 50 \\
\#Scenes & 117{,}503 & 8{,}052 & 8{,}156 \\
\#Salient scenes & 16{,}678 & 2{,}677 & 2{,}688 \\
Salient rate (\%) & 14.19 & 33.25 & 32.96 \\
\bottomrule
\end{tabular}
\caption{MENSA split statistics highlighting the distribution shift between silver training labels and gold evaluation labels.}
\label{tab:mensa_split_stats}
\end{table}

\begin{figure}[htbp]
  \centering
  \includegraphics[
  width=0.9\linewidth,
  height=10cm,
  keepaspectratio,
  trim=2.5cm 10cm 2.5cm 10cm,
  clip
]{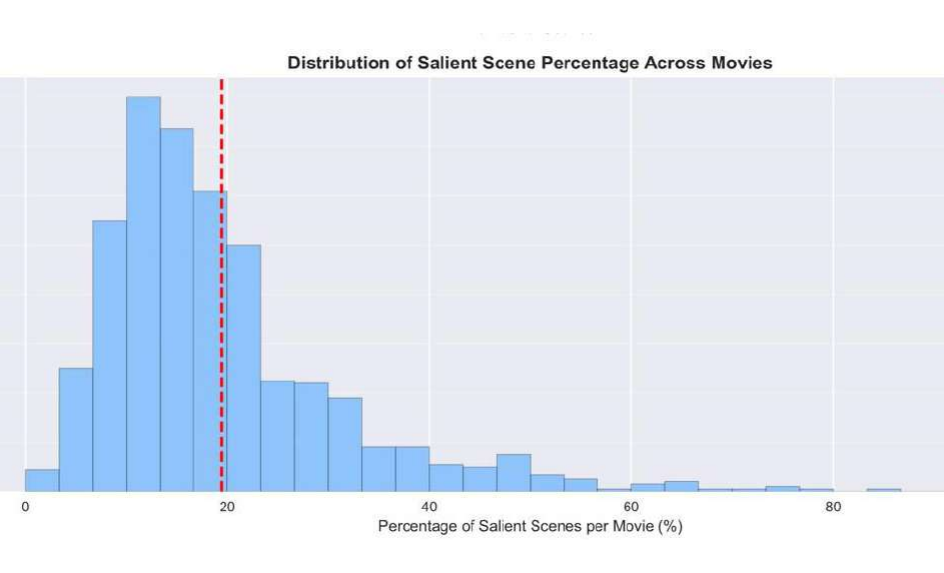}

  \caption{Distribution of salient scene percentage across movies.}
  \label{fig:mensa_salient_pct}
\end{figure}

\paragraph{Imbalance and Biases.}
MENSA is highly imbalanced: only 16.49\% of scenes are salient overall, with a pronounced label distribution shift between training (14.19\%) and validation/test ($\approx$33\%). This makes accuracy unreliable and motivates imbalance-aware objectives. Additionally, salient scenes are significantly longer (mean 355.7 vs.\ 132.8 words), introducing a strong verbosity bias, and are slightly more frequent in later parts of scripts (18.92\%), indicating mild positional bias (Figure~\ref{fig:mensa_salient_pct}).

Overall, MENSA poses a challenging setting due to noisy supervision, severe class imbalance, and structural biases, making it a rigorous benchmark for robust scene salience modeling.

\section{Computational Complexity and Pipeline Latency}
\label{app:computational_complexity}

LGSM integrates several high-capacity upstream extractors (e.g., GPT-2, KenLM, and neural RST parsers). To eliminate runtime overhead, we decouple feature extraction from sequence modeling. By pre-computing all linguistic features and frozen BERT-large embeddings offline, we ensure the online optimization phase remains highly efficient. As shown in Table \ref{tab:training_latency}, the architecture converges rapidly on standard hardware, with the sequence modeling component maintaining a negligible computational footprint.

Regarding pipeline reliability, this architecture is theoretically susceptible to cascading errors from upstream components. However, the Adaptive Gated Fusion mechanism acts as a robust filter; it effectively learns to suppress noisy linguistic signals, prioritizing the stable semantic representations provided by the frozen BERT-large encoder when upstream structural features are anomalous.

\begin{table}[h]
\centering
\small
\setlength{\tabcolsep}{4pt}
\begin{tabular*}{\columnwidth}{l @{\extracolsep{\fill}} r}
\toprule
\textbf{Metric} & \textbf{Specification} \\
\midrule
Hardware & 1$\times$ NVIDIA Quadro RTX 6000 \\
CPU & Dual AMD EPYC 9124 (3.0 GHz) \\
Training Corpus & 824 Movies (117,503 Scenes) \\
LGSM input dimension & 888 \\
Total Epochs & 25 \\
Batch Size & 4 \\
Time per Epoch & $\sim$6 to 7 seconds \\
Convergence Time & $<$ 3 minutes \\
\bottomrule
\end{tabular*}
\caption{Empirical training profile for the LGSM architecture.}
\label{tab:training_latency}
\end{table}

\section{Error Analysis}
\label{app:error_analysis}

\subsection{Error decomposition and false-positive signature}

\begin{figure}[htbp]
    \centering
    \includegraphics[width=\linewidth]{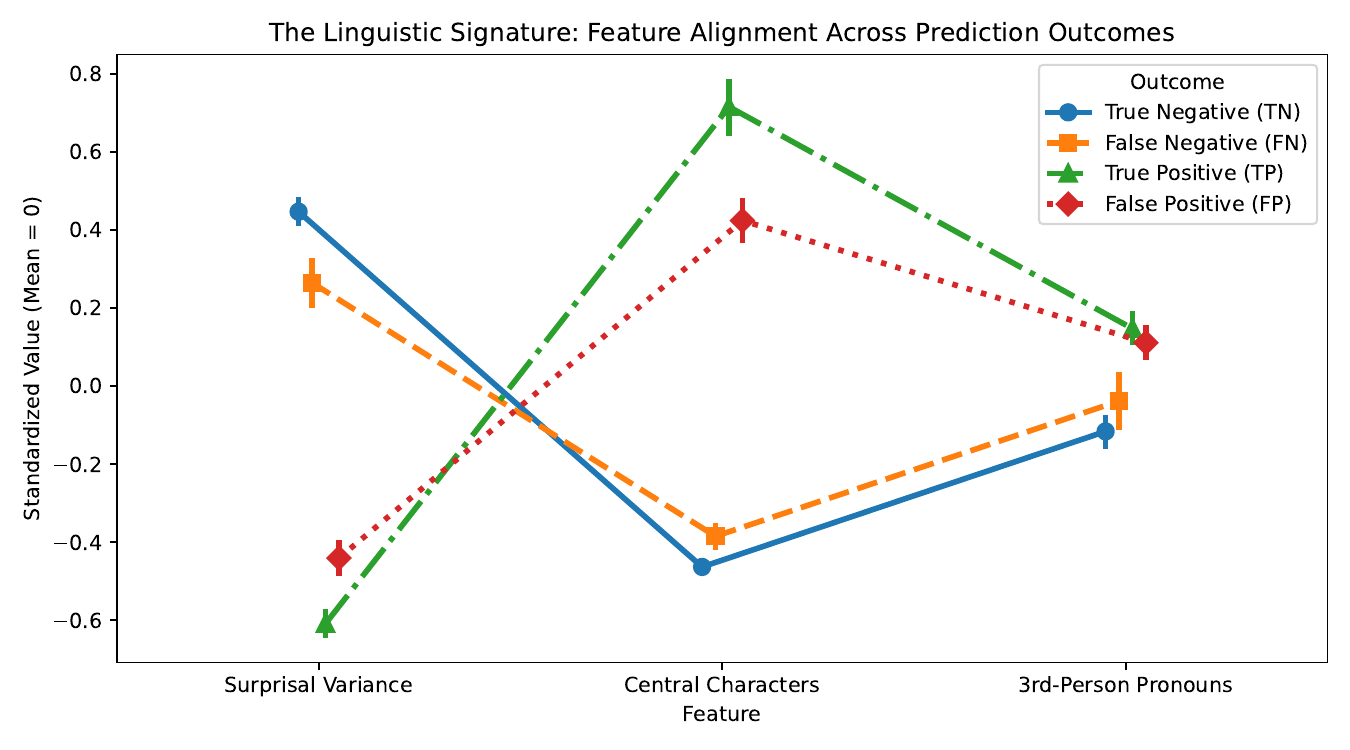}
    \caption{Standardised feature profiles across prediction outcomes. False Positives (FPs) structurally mirror True Positives (TPs).}
    \label{fig:feature_alignment}
\end{figure}

Table~\ref{tab:error_comparison} shows the raw prediction counts on the MENSA test set. LGSM recovers 451 salient scenes missed by the semantic baseline (true positives: 1,837 vs.\ 1,386) but also predicts 736 additional false positives (1,702 vs.\ 966). This divergence partly reflects the limitations of the silver-standard training data: MENSA's ground truth is derived by aligning scripts to compressed Wikipedia plot summaries, so scenes are labelled non-salient if they are omitted from the synopsis. Because LGSM relies on intrinsic structural cues (e.g., character convergence), it tends to flag narratively significant setup or tension scenes that exist in the structure but are too granular for the high-level summary. Figure~\ref{fig:feature_alignment} supports this interpretation: false positives structurally mirror true positives along the strongest linguistic features (BERT surprisal variance, central character count, third-person pronoun count), rather than clustering with the true negatives, which is consistent with FPs being scenes that look salient on linguistic grounds and are penalised primarily by the silver labelling protocol. The downstream summarisation evaluation (Appendix~\ref{app:summarization_results}) supports this interpretation: including these structurally identified scenes improves generation quality (\textbf{+3.67} ROUGE-1, \textbf{+1.51} ROUGE-L) without destabilising BERTScore.

\begin{table}[h]
\centering
\small
\setlength{\tabcolsep}{3pt}
\begin{tabular}{@{}l|cc|cc@{}}
\toprule
\textbf{Model} & \textbf{TP (Recall)} & \textbf{FN (Missed)} & \textbf{FP (Noise)} & \textbf{TN} \\
\midrule
Baseline & 1,386 (51.6\%) & 1,302 & \textbf{966} & \textbf{4,502} \\
\textbf{LGSM} & \textbf{1,837 (68.3\%)} & \textbf{851} & 1,702 & 3,766 \\
\bottomrule
\end{tabular}
\caption{Error decomposition on the Mensa test set.}
\label{tab:error_comparison}
\end{table}

\subsection{False-negative outlier taxonomy}

We categorize false-negative errors using highly influential linguistic features as proxies, selected based on their global influence measured by mean absolute SHAP values. For each feature, we compute the proportion of false-negative scenes (restricted to the subset for which SHAP explanations are available) in which the feature exhibits an extreme value (at least $3\sigma$ from the norm). Table~\ref{tab:error-proxy-fn} presents the resulting taxonomy. This analysis suggests that despite substantial recall gains, errors arise from extreme linguistic conditions rather than missing cues. Specifically, the model consistently misses scenes with abnormal surprisal or positional patterns, indicating that while it captures standard narrative logic, it degrades in atypical regimes where multiple strong cues interact or deviate from the training distribution.

\begin{table}[htbp]
\centering
\small
\begin{tabular}{l c}
\toprule
\textbf{Error Proxy (Feature Outlier)} & \textbf{\% of FN Scenes} \\
\midrule
BERT positional deviation & 1.04 \\
Minimum surprisal  & 0.52 \\
Graph-based concept coverage  & 0.52 \\
Surprisal variability  & 0.26 \\
\bottomrule
\end{tabular}
\caption{Distribution of false-negative errors of LGSM model linked to specific feature outliers ($>3\sigma$).}
\label{tab:error-proxy-fn}
\end{table}

\section{Downstream Summarization}
\label{app:summarization_results}

To validate extraction utility, we fine-tune an \textbf{LED-Large} summarizer \citep{Beltagy2020} on scenes extracted by the baseline versus LGSM.

\begin{table*}[htbp]
\centering
\small
\setlength{\tabcolsep}{4.5pt}
\begin{tabular}{l|ccc|c}
\toprule
\textbf{Extraction Source} & \textbf{R-1} & \textbf{R-2} & \textbf{R-L} & \textbf{BERT-F1} \\
\midrule
Select and Summarize (SOTA) & 38.02 & \textbf{9.15} & 15.69 & \textbf{83.29} \\
\textbf{LGSM (Ours)} & \textbf{41.69} & 8.88 & \textbf{17.20} & 83.17 \\
\midrule
\textit{Improvement} & \textit{\textbf{+3.67}} & \textit{-0.27} & \textit{\textbf{+1.51}} & \textit{-0.12} \\
\bottomrule
\end{tabular}
\caption{LGSM improves content coverage (ROUGE-1/L) with negligible semantic loss (BERT-F1).}
\label{tab:summarization_results}
\end{table*}

Table~\ref{tab:summarization_results} shows that LGSM drives substantial gains in ROUGE-1 (+3.67) and ROUGE-L (+1.51) compared to the baseline. By passing 1,187 more scenes to the summariser (3,539 vs.\ 2,352 predicted-salient scenes), LGSM provides the structural setup material required for comprehensive narrative generation, whereas the baseline effectively starves the model of context.

Despite a marginal dip in exact bigram matching (ROUGE-2 $\downarrow$ 0.27), BERTScore remains statistically equivalent (83.17 vs.\ 83.29). This confirms that while LGSM-based summaries may vary in phrasing, they maintain the baseline's semantic fidelity, effectively trading filterable redundancy for narrative completeness.

\section{SHAP Contribution Patterns}
\label{app:shap_contributions}

\begin{figure}[htbp]
    \centering
    \includegraphics[width=0.95\linewidth]{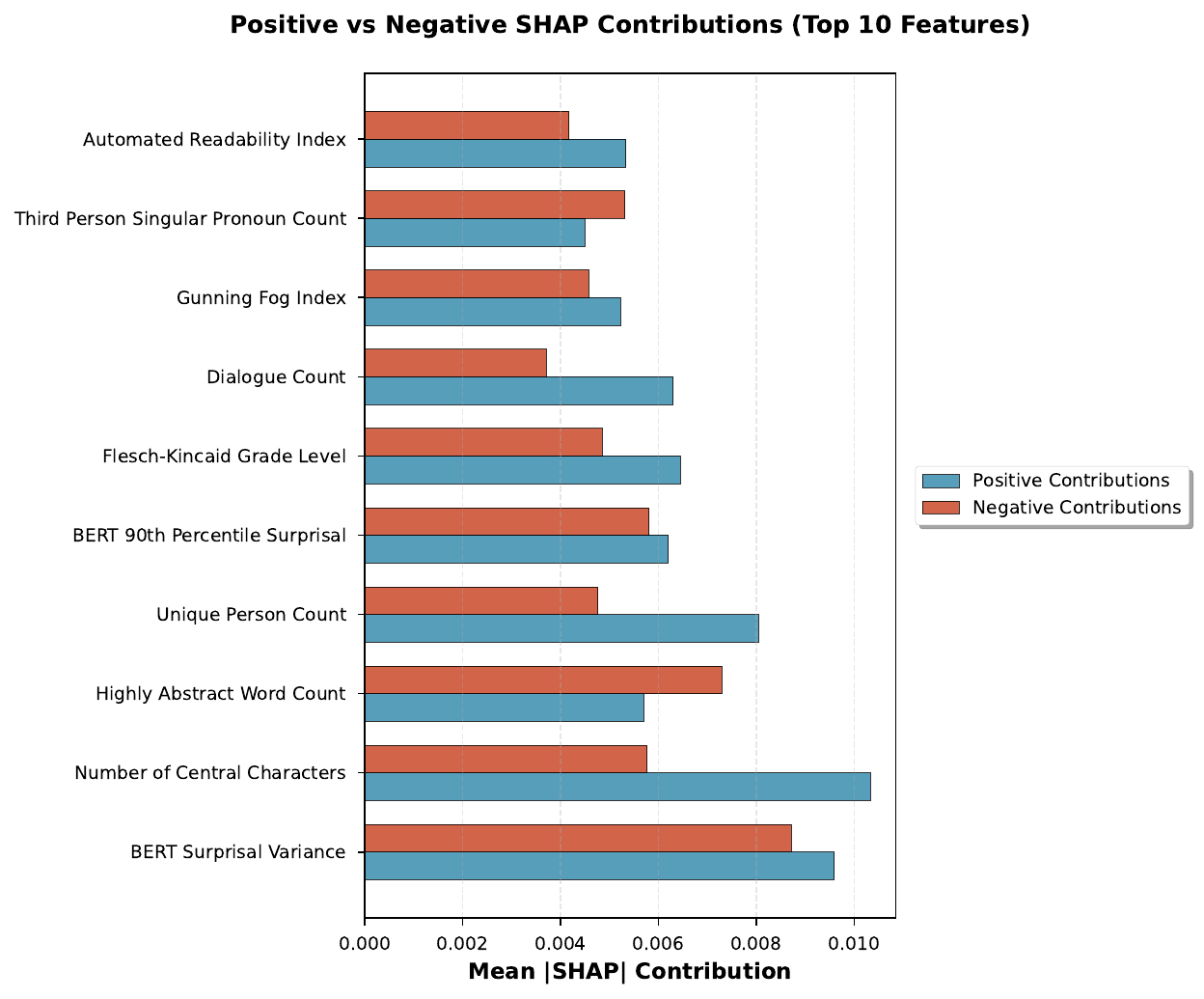}
    \caption{Mean positive and negative SHAP contributions for the top ten features, indicating conditional feature effects.}
    \label{fig:shap_contributions_appendix}
\end{figure}

Figure~\ref{fig:shap_contributions_appendix} presents the mean positive and negative SHAP contributions for the top ten features. The bidirectional pattern confirms conditional feature usage: features such as readability, dialogue count, and abstraction contribute both positively and negatively to salience predictions depending on context. This supports the interpretation in Section~\ref{sec:shap_analysis} that LGSM evaluates these features relationally rather than through simple monotonic heuristics.

\section{Token-Level Surprisal Distributions}
\label{app:token_level_distributions}

\begin{figure*}[htbp]
    \centering
    \includegraphics[width=\textwidth]{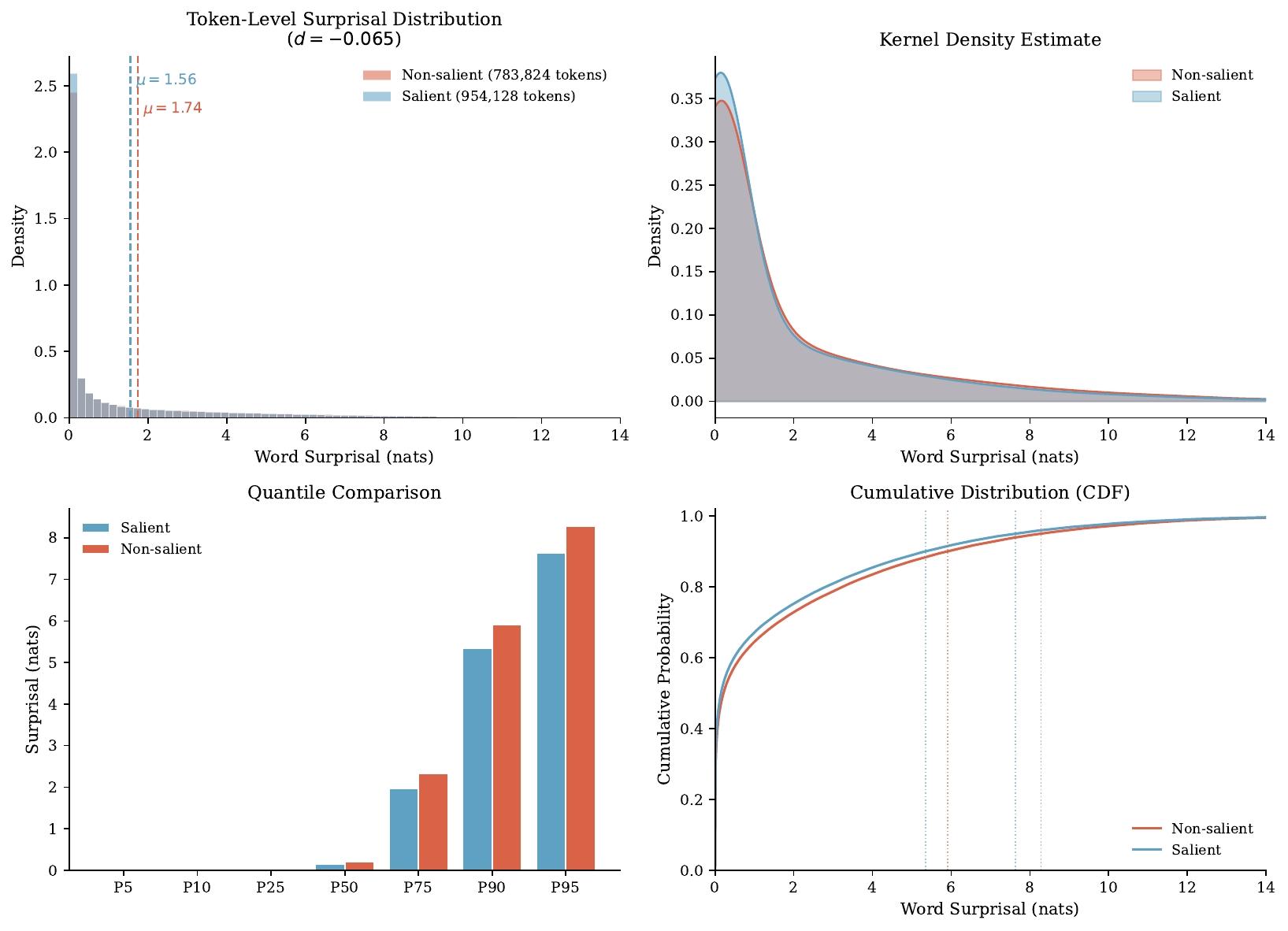}
    \caption{Token-level BERT surprisal distributions for salient and non-salient tokens on the MENSA validation set. \textbf{Top-left:} Overlaid histograms with mean markers indicating near-complete distributional overlap (Cohen's $d = -0.065$). \textbf{Top-right:} Kernel density estimates confirming equivalent distributional shapes. \textbf{Bottom-left:} Quantile comparison showing negligible differences below the median and meaningful divergence at $P_{90}$ ($\Delta = -0.56$) and $P_{95}$ ($\Delta = -0.66$). \textbf{Bottom-right:} Cumulative distribution functions; the curves track closely throughout, with the salient CDF rising consistently faster above the median and a $0.56$-nat horizontal gap at the 90th percentile.}
    \label{fig:token_level_full}
\end{figure*}

This appendix presents the complete distributional comparison of word-level BERT surprisal values between salient and non-salient tokens on the MENSA validation split, supplementing the cumulative distribution analysis reported in the Discussion. Figure~\ref{fig:token_level_full} provides four complementary views of the data ($n_{\text{salient}} = 954{,}128$ tokens; $n_{\text{non-salient}} = 783{,}824$ tokens; the salient class has more tokens despite fewer scenes because salient scenes are roughly three times longer on average).

Table~\ref{tab:token_level_stats} summarizes the key distributional statistics on both gold-labelled splits. The near-identical distributions (histogram overlap $= 97.1\%$ on val; $96.7\%$ on test) indicate that salient and non-salient scenes draw from effectively the same token-level surprisal distribution, with divergence restricted to the suppression of extreme upper-tail values in salient text.

\begin{table}[t]
\centering
\footnotesize
\setlength{\tabcolsep}{5pt}
\begin{tabular}{l rr rr}
\toprule
& \multicolumn{2}{c}{\textbf{Val}} & \multicolumn{2}{c}{\textbf{Test}} \\
\cmidrule(lr){2-3}\cmidrule(lr){4-5}
\textbf{Statistic} & \textbf{Sal.} & \textbf{Non-sal.} & \textbf{Sal.} & \textbf{Non-sal.} \\
\midrule
Mean (nats)       & $1.56$ & $1.74$ & $1.62$ & $1.78$ \\
Median (nats)     & $0.15$ & $0.21$ & $0.14$ & $0.21$ \\
Std (nats)        & $2.74$ & $2.91$ & $3.18$ & $3.21$ \\
Skewness          & $2.59$ & $2.29$ & $4.16$ & $3.51$ \\
Kurtosis          & $9.56$ & $6.30$ & $29.11$ & $22.03$ \\
\midrule
& \multicolumn{2}{c}{\textbf{Diff. (val)}} & \multicolumn{2}{c}{\textbf{Diff. (test)}} \\
\cmidrule(lr){2-3}\cmidrule(lr){4-5}
$\Delta$\,mean    & \multicolumn{2}{c}{$-0.18$}   & \multicolumn{2}{c}{$-0.16$} \\
$\Delta$\,median  & \multicolumn{2}{c}{$-0.063$}  & \multicolumn{2}{c}{$-0.067$} \\
$\Delta P_{5}$    & \multicolumn{2}{c}{${\approx}0$}  & \multicolumn{2}{c}{${\approx}0$} \\
$\Delta P_{10}$   & \multicolumn{2}{c}{${\approx}0$}  & \multicolumn{2}{c}{${\approx}0$} \\
$\Delta P_{25}$   & \multicolumn{2}{c}{${-}0.002$}    & \multicolumn{2}{c}{${-}0.002$} \\
$\Delta P_{50}$   & \multicolumn{2}{c}{$-0.063$}  & \multicolumn{2}{c}{$-0.067$} \\
$\Delta P_{75}$   & \multicolumn{2}{c}{$-0.372$}  & \multicolumn{2}{c}{$-0.392$} \\
$\Delta P_{90}$   & \multicolumn{2}{c}{$-0.558$}  & \multicolumn{2}{c}{$-0.531$} \\
$\Delta P_{95}$   & \multicolumn{2}{c}{$-0.655$}  & \multicolumn{2}{c}{$-0.613$} \\
\midrule
Cohen's $d$       & \multicolumn{2}{c}{$-0.065$}  & \multicolumn{2}{c}{$-0.052$} \\
Histogram overlap & \multicolumn{2}{c}{$0.971$}   & \multicolumn{2}{c}{$0.967$} \\
KS $D$            & \multicolumn{2}{c}{$0.029$}   & \multicolumn{2}{c}{$0.032$} \\
\bottomrule
\end{tabular}
\caption{Token-level surprisal statistics on the MENSA validation and test splits, computed over all word-level BERT surprisals (val: $1{,}737{,}952$ tokens; test: $1{,}706{,}362$ tokens). All distributional differences are statistically significant ($p < 0.001$) but of negligible practical magnitude in the bulk, with divergence concentrated in the upper quantiles ($P_{90}, P_{95}$).}
\label{tab:token_level_stats}
\end{table}

\end{document}